\pdfoutput=1

\documentclass[11pt]{article}

\usepackage[final]{acl}

\usepackage{times}
\usepackage{latexsym}

\usepackage[T1]{fontenc}

\usepackage[utf8]{inputenc}

\usepackage{microtype}

\usepackage{inconsolata}

\usepackage{array}
\usepackage{pifont}
\usepackage{tabularx}
\usepackage{adjustbox}
\usepackage{multirow}
\usepackage{enumitem}
\usepackage{xspace}
\usepackage{tcolorbox}
\usepackage{booktabs,amsfonts,dcolumn}
\usepackage{hyperref}
\usepackage{url}
\usepackage{amsmath,amsthm,amsfonts,amssymb,bm,stmaryrd,bbm}
\usepackage[noorphans,vskip=0.75ex,leftmargin=2ex]{quoting}
\usepackage{footmisc}\interfootnotelinepenalty=10000
\usepackage{colortbl}
\usepackage{cleveref}
\providecommand{\todo}[1]{{\protect\color{cyan}{[TODO: #1]}}}
\providecommand{\danqi}[1]{{\protect\color{orange}{[Danqi: #1]}}}
\providecommand{\tianyu}[1]{{\protect\color{blue}{[Tianyu: #1]}}}
\providecommand{\howard}[1]{{\protect\color{purple}{[Howard: #1]}}}
\providecommand{\revision}[1]{{#1}}

\providecommand{\todo}[1]{{\protect\color{red}{}}}
\providecommand{\danqi}[1]{{\protect\color{orange}{}}}
\providecommand{\tianyu}[1]{{\protect\color{blue}{}}}
\providecommand{\howard}[1]{{\protect\color{purple}{}}}

\newcommand\ti[1]{\textit{#1}}

\newcommand\tf[1]{\textbf{#1}}
\newcommand\ttt[1]{\texttt{#1}}

\renewcommand{\paragraph}[1]{\vspace{0.2cm}\noindent\textbf{#1}}

\newcommand{\replug}{{\sc{RePlug}}}

\newcommand{\retdoc}{RetDoc}

\newcommand{\ours}{{\sc{CEPE}}}
\newcommand{\ourschat}{{\sc{CEPED}}}
\newcommand{\oursllama}{{{\sc{CEPE-LLaMA-2}}}}
\newcommand{\oursllamachat}{{{\sc{CEPED-LLaMA-2-Chat}}}}

\newcommand{\llama}{{\sc{LLaMA-2}}}
\newcommand{\llamachat}{{\sc{LLaMA-2-Chat}}}
\newcommand{\llamak}{{\sc{LLaMA-2-32K}}}
\newcommand{\yarn}{{\sc{YaRN-64K}}}
\newcommand{\yarnmore}{{\sc{YaRN-128K}}}
\newcommand{\llamakchat}{{\sc{LLaMA-2-32K Instruct}}}
\newcommand{\streaming}{{\sc{StreamingLLM}}}
\newcommand{\retro}{{\sc{Retro}}}

\newcommand{\zs}{Zero{\sc{Scrolls}}}

\newcommand{\rpfilter}{RP$_\text{train-filter}$}
\newcommand{\rpcat}{RP$_\text{train-cat}$}

\newcommand{\menc}{\mathcal{M}_\textrm{enc}}
\newcommand{\mdec}{\mathcal{M}_\textrm{dec}}
\newcommand{\concat}{\textsc{concat}}

\newcommand{\denc}{d_\textrm{enc}}
\newcommand{\ddec}{d_\textrm{dec}}

\NewDocumentCommand\logo{}{\includegraphics[width=0.35cm]{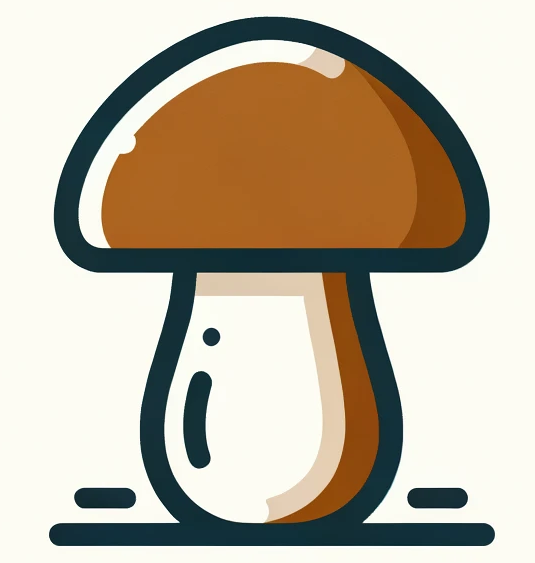}}

\title{Long-Context Language Modeling with Parallel Context Encoding}

\author{Howard Yen \quad Tianyu Gao \quad Danqi Chen  \\
Princeton Language and Intelligence (PLI), Princeton University \\
\ttt{\{hyen,tianyug,danqic\}@cs.princeton.edu}
}

\begin{document}
\maketitle

\begin{abstract}

Extending large language models (LLMs) to process longer inputs is crucial for a wide range of applications.
However, the substantial computational cost of transformers and limited generalization of positional encoding restrict the size of their context window.
We introduce \tf{C}ontext \tf{E}xpansion with \tf{P}arallel \tf{E}ncoding (\tf{\ours} {\logo{}}),
a framework that 
can be applied to 
any existing decoder-only LLMs to extend their context window. %
CEPE employs a small encoder to process long inputs chunk by chunk, enabling the frozen decoder to utilize additional contexts via cross-attention. CEPE is efficient, generalizable, and versatile: trained with 8K-token documents, it extends the context window of LLAMA-2 to 128K tokens, offering 10× the throughput with only 1/6 of the memory. 
\ours{} yields strong performance on language modeling and in-context learning.
\ours{} also excels in retrieval-augmented applications,
while existing long-context models degenerate with retrieved contexts.
We further introduce a \ours{} variant that can extend the context window of instruction-tuned models using only unlabeled data, and showcase its effectiveness on \llamachat{}, leading to a strong instruction-following model that can leverage very long contexts on downstream tasks.\footnote{Code and models are available at~\url{https://github.com/princeton-nlp/CEPE}.}

\end{abstract}

\begin{figure}[t!]
    \centering
    \includegraphics[width=0.98\linewidth]{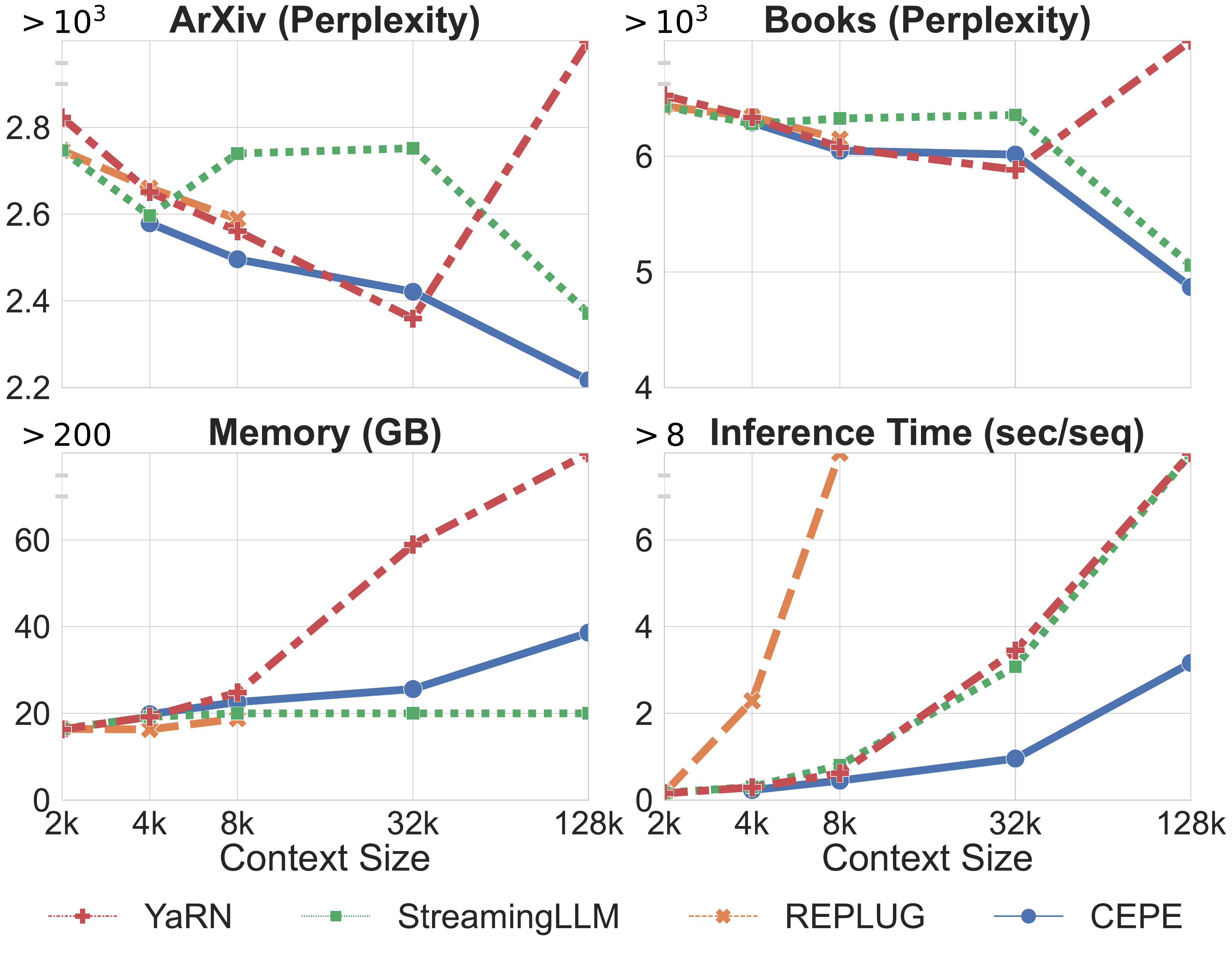}
    \caption{
        A comparison between \ours{} 
        and other techniques of extending LLMs' context window, including {\sc{YaRN}} \cite{peng2024yarn}, \streaming{} \cite{xiao2024efficient}, and \replug{} \cite{shi2023replug}.
        \ours{} trained on 8K tokens can generalize to 128K tokens with minimal computational and memory costs.
    }
    \label{fig:teaser}
\end{figure}

\section{Introduction}
Enabling long and extensible context is crucial 
for large language models (LLMs) to effectively perform complex tasks,
such as summarizing a book or answering questions with hundreds of retrieved Web pages. 
However, several challenges limit the ability of LLMs to leverage long context:
(1) LLMs and popular positional encodings \cite{raffel2020exploring,su2021roformer}
do not generalize to sequence lengths longer than the lengths seen during training \cite{press2022train}, even after additional fine-tuning \cite[][\emph{inter alia}]{chen2023extending,chen2024longlora,peng2024yarn}.
(2) Transformers~\cite{vaswani2017attention}---the predominant architecture of LLMs---incur a quadratic computational cost and a linear memory cost with respect to the input length,
making it expensive to use for long sequences.
(3) High-quality long-context data, 
such as long instruction-following data, are scarce and difficult to obtain \cite{wang-etal-2023-self-instruct,xiong2023effective}.

Besides directly fine-tuning LLMs on longer inputs, %
a series of inference-time modification methods have been proposed recently to 
scale up the effective context window, %
either by modifying the attention mechanism \cite{bertsch2023unlimiformer,xiao2024efficient,ivgi2023sled} 
or encoding chunks of context in separate forward passes \cite{shi2023replug,ratner-etal-2023-parallel,lin2023radit}. %
While these methods generalize to longer sequences,
the model often fails to effectively leverage the extra tokens 
and can incur greater inference costs.

\begin{figure*}[t!]
    \centering
    \includegraphics[width=0.90\linewidth]{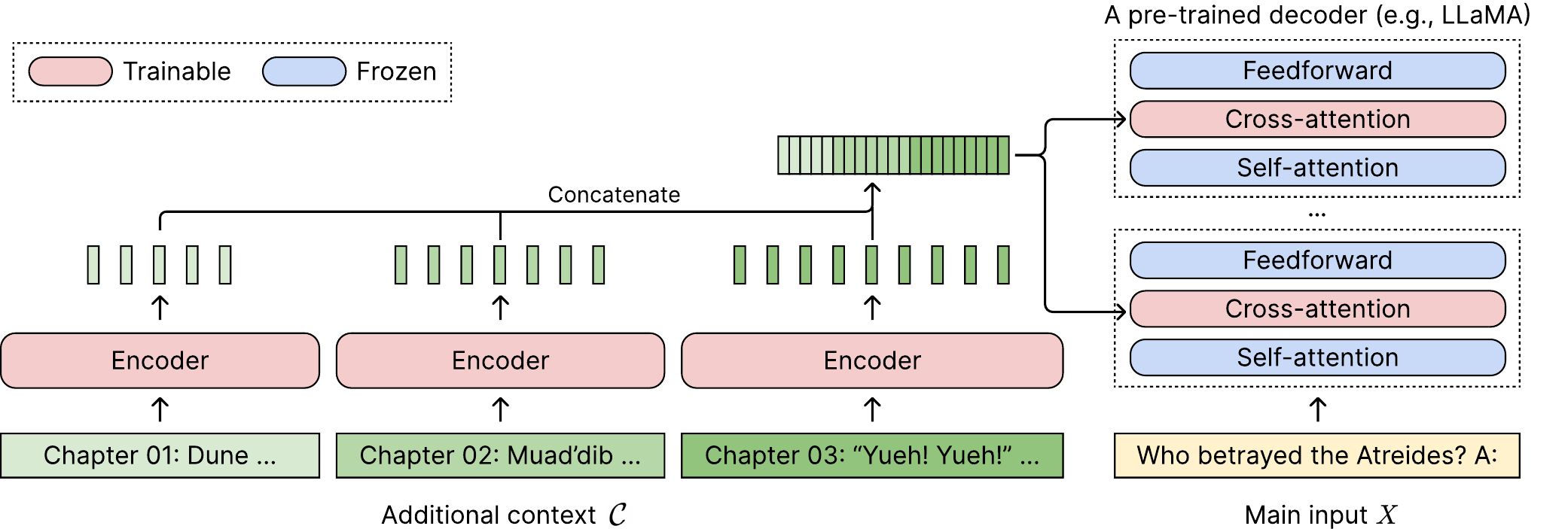}
    \caption{
        The \ours{} {\logo{}} architecture. 
        The encoder model encodes the additional 3 chunks ($k=3$) of context $\mathcal{C}$ 
        in parallel, and the final hidden representations from the encoder model are concatenated and used as inputs to the cross-attention layers in the decoder model.
        The cross-attention layers attend to the encoder representations between the self-attention and feed-forward layers in the decoder model.
    }
    \vspace{-3pt}
    \label{fig:architecture}
\end{figure*}

In this work, we propose an efficient and lightweight solution to extending the context window of LLMs, called \tf{C}ontext \tf{E}xpansion with \tf{P}arallel \tf{E}ncoding (\tf{\ours{}} {\logo{}}). 
\ours{} is applicable to any
pre-trained decoder-only LM by adding two components:
(1) a small encoder that encodes the long context in chunks, and
(2) a cross-attention module that is inserted at each layer of the decoder to attend to the encoder representations (\Cref{fig:architecture}).
With a careful selection of unlabeled training data, {\ours} can leverage not only long-context documents but also retrieved documents effectively. %

\ours{} offers several benefits:
(1) \tf{length generalization}: 
\ours{} is not limited by positional encoding constraints 
as the long context is encoded in chunks, each with its own positional encoding;
(2) \tf{efficiency}:
using a small encoder and processing contexts in parallel reduce computational cost.
Since cross-attention attends only to the last layer's representations from the encoder,
\ours{} requires much less memory compared to decoder-only LMs,
which cache the key-value pairs of every token in every layer;
(3) \tf{reduced training cost}: 
unlike full fine-tuning approaches, 
we only tune the encoder and the cross-attention while keeping the large decoder LM frozen; 
augmenting a 7B decoder with a 400M encoder and cross-attention layers (1.4B parameters) can be done with one A100 80GB GPU.

We apply \ours{} to \llama{}~\cite{touvron2023llama2} and train it on a filtered version of RedPajama~\citep{together2023redpajama} for 20B tokens---only $1$\% 
of the pre-training budget of \llama{}. 
We first show that \oursllama{}, trained with 8K input length, continues to improve perplexity on longer input up to 128K tokens. 
Then, we apply \ours{} to 
a retrieval-augmented setting,
as our larger context window allows incorporating more retrieved documents.
Compared to existing 
methods,
\ours{} achieves better performance on both retrieval-augmented language modeling 
and open-domain question answering.
Additionally, we demonstrate that \ours{} can effectively leverage more demonstrations for in-context learning~\cite{brown2020language}.
All the above is achieved with a much lower memory and computational cost than most previous solutions.

Finally, we propose \ours-{{\textbf{D}istilled}} (\ourschat{}), which extends the context window of \ti{instruction-tuned} models, using only unlabeled data.
\ourschat{} distills the behavior of the original instruction-tuned model to the new architecture through an auxiliary KL divergence loss,
which eliminates the need to curate expensive long-context instruction-following data~\cite{ivison2023camels}.
We apply \ourschat{} to \llamachat{} and show that while preserving their instruction understanding ability,
\oursllamachat{} can incorporate more context and improve performance on long-text understanding tasks~ \citep{shaham-etal-2023-zeroscrolls}.

To conclude,
\ours{} is a lightweight framework that can extend context windows of 
any base or instruction-tuned LMs.
We hope \ours{} can empower future LLM research with 
cheap and effective long-context abilities.

\section{Method: \ours{} {\logo{}}}

\label{sec:methods}
We design \ours{} to adapt pre-trained LLMs to perform long-context language modeling on sequences with many tokens (e.g., books).
For retrieval augmentation, these long contexts may contain a set of retrieved passages instead.
We first describe how \ours{} modifies the architecture of LLMs to attend to representations encoded by a small encoder, and then describe how the \ours{} modules are trained.
Finally, we extend \ours{} to \ourschat{}, which expands the context window of instruction-tuned models using only unlabeled data.

\subsection{Architecture}\label{sec:architecture}

\ours{} augments off-the-shelf decoder-only LMs by %
(1) adding a small, bidirectional pre-trained encoder model and 
(2) inserting cross-attention layers between the self-attention and feed-forward layers in every transformer block of the decoder model.

\paragraph{Notation.}
Given an input context with $T$ tokens $x_1,...,x_T$, we consider the first $m$ tokens $x_1,...,x_m$ as the additional context $\mathcal{C}$ and the last $n = T - m $ 
tokens $x_{m+1},...,x_T$ as the main input $X$.
The additional context 
is split into chunks 
$\mathcal{C} = C_1,...,C_k$, which can contain either segments within a long document or a set of retrieved passages. %
We use $\menc$ to denote the encoder model with hidden dimension $\denc$ and $\mdec$ to denote the decoder-only LLM with hidden dimension $\ddec$.

\paragraph{Encoding chunks.}
We first encode $C_1,...,C_k$ chunk by chunk using the trainable encoder $\menc$: 
\begin{equation*}
\phi_i = \menc(C_i) \qquad \Phi = \concat(\{\phi_i\}_{i=1}^k) 
\end{equation*}
where $\phi_i \in\mathbb{R}^{|C_i|\times \ddec}$ is the token-wise last layer hidden state from $\menc$ and $\Phi\in\mathbb{R}^{m\times \ddec}$.
Note that $\menc$ is bidirectional, which results in more information-rich representations compared to unidirectional ones. 
\revision{While we do not preserve global positions across different chunks, experiments show that \ours{} achieves better or comparable performance to full-attention models that do.}

\paragraph{Cross-attention modules.}
In every decoder layer of the transformer, we insert a cross-attention module between the self-attention and feed-forward layers. %
To construct the cross-attention module, we provide $\Phi$ as keys and values, and the hidden states of $X$ as queries.
Note that in order for $\mdec$ to attend to $\Phi$, the key and value projection matrices in the cross-attention module also serve as an up-projection that transforms the $\denc$-dimensional $\Phi$ into a $\ddec$-dimensional embedding.
\Cref{fig:architecture} illustrates the architecture of \ours{}.

\paragraph{Efficiency.}
$\menc$ 
is much smaller 
and 
encodes contexts in parallel to avoid the quadratic complexity of full attention.
This enables \ours{} to exhibit a substantially higher training and inference speed
than if we were to use $\mdec$ to process all $T = m + n$ tokens.
Moreover, \ours{} drastically reduces memory consumption by avoiding caching $(m+n)L$ key-value pairs ($L$ is the number of layers of $\mdec$) and instead caching only $\Phi$ and $nL$ key-value pairs.
In particular, using a standard decoder-only model requires $\mathcal{O}((m+n)L\ddec)$ memory whereas \ours{} requires $\mathcal{O}(m\denc + nL\ddec)$. 
In our setting, $m\gg n$ and $\ddec \gg \denc$, so in practice, we observe a substantial gap: for each additional token, \ours{} requires only 1/256 of the memory compared to encoding them in $\mdec$. 

\paragraph{Comparison with existing retrieval-augmented LMs.}
\ours{} takes inspiration from retrieval-augmented models such as 
FiD \cite{izacard2021leveraging}, Atlas \cite{izacard2022atlas}, and \retro{} \cite{borgeaud2022retro}
for attending to parallel encodings. 
Our method differs in the following key aspects: 
FiD, Atlas, and \retro{} are all full-parameter training methods for encoder-decoder models; 
Atlas and \retro{} are pre-training methods that rely on expensive large-scale retrieval-augmented data;
FiD is a fine-tuning method for open-domain QA tasks, which requires task-specific data.
In contrast, \ours{} is a lightweight framework that can extend the context window of any existing decoder-only model; 
we only fine-tune an added small encoder and the cross-attention layers;
instead of retrieval-augmented data,
\ours{} only requires
a small amount of filtered unlabeled data, as described in the following section;
because \ours{} is built on top of powerful pre-trained models,
it can be easily adapted to a wide range of tasks without explicit fine-tuning.

\subsection{Data}
\label{sec:methods:data}

We use RedPajama \cite[RP;][]{together2023redpajama} as our training corpus, which 
is an open-source reproduction of the {\sc{LLaMA}}-1 \cite{touvron2023llama} training data.
It contains about 1 trillion tokens from seven domains: ArXiv, Books, C4-RP, CC, Github, StackExchange, and Wiki.
We first partition the corpus into three sets of documents: training, test, and retrieval;
we leverage the retrieval corpus for retrieval-augmented language modeling.

We design a data mixture with strong long-range dependencies and a diversity of domains.
The training data is preprocessed into two subsets,
each of which is a collection of $8192$-token sequences:
(1) In \rpcat{}, we concatenate documents together to form training sequences;
(2) In \rpfilter{}, we keep documents from the Arxiv and Books domains 
that have at least $8,192$ tokens and sample sequences within document boundaries. 

Our qualitative analysis found that data from the ArXiv and Books domains naturally contain long documents that are especially useful when training long-context models.
It is also important to use a mixture of data from all domains 
to ensure better generalization.
Thus, %
we use a mixture ratio of 2:1 between 
\rpfilter{} and \rpcat{}
for training.
We alternatively consider using retrieval supervision when generating the training data to improve the model's ability to leverage retrieved documents.
However, our ablations found that our unlabeled data mixture transfers well to retrieval-augmented settings at a much cheaper cost (\S\ref{sec:ablations:training}).

\subsection{Training}
\label{sec:methods:training}
We use \llama-7B \cite{touvron2023llama2} as $\mdec$ (originally trained on 4K length), and insert the new cross-attention layers described in \Cref{sec:architecture} into it.
We add a bidirectional encoder $\menc$ with 435M parameters, yielding $1.8$B added parameters in the \ours{} model.

\paragraph{Encoder.}
We first pre-train a bidirectional masked-language model %
on the RedPajama dataset.
$\menc$ follows the configuration of RoBERTa-large \cite{liu2019roberta} but shares the vocabulary with $\mdec$, \llama.
\footnote{Future works applying \ours{} to \llama{} models can simply use our pre-trained encoder.}
We train $\menc$ for 100K steps with a batch size of $2,048$ and sequence length of $512$ tokens. For more details, refer to \S\ref{app:training_encoder}.

\paragraph{Cross-attention.}
We freeze the weights of $\mdec$ and train only the added cross-attention layers as well as fine-tuning 
$\menc$ using the cross-entropy loss.
We first adopt a warmup training stage designed to teach $\mdec$ to copy from $\menc$ through cross-attention---for each position $i \leq T$, the objective is to generate $x_{i+1}$, conditioned on $\menc(x_1,\ldots,x_T)$ and $\mdec(x_1,\ldots,x_i)$---$\menc$ and $\mdec$ share the same input.
The warmup stage uses 131M tokens from the training set of RP.\footnote{Preliminary experiments suggest this stage can stabilize training; ablations can be found in \S\ref{sec:ablations:training}.}

After the warmup stage, we move to the standard training, where  
each sequence has $T=8,192$ tokens. 
We use the last $4,096$ tokens as the decoder input $X$, and chunk the first $4,096$ tokens into $k=16$ contexts of $|C_i| = 256$ tokens each as the encoder input $\mathcal{C}$, and train for 20B tokens.
Freezing the decoder allows \ours{} to be trained on a single A100 GPU, which is a significant reduction in computational cost compared to training $\mdec$ with sequence length $T$.
For more training details and hyperparameters, please refer to \S\ref{app:training}.

\subsection{\ourschat{} for Instruction-Tuned Models}
We extend our method to \ours-{\sc{\textbf{D}istilled}} (\ourschat{})
to augment instruction-tuned models with longer context.
Instruction-tuned models \cite{ouyang2022training,alpaca,touvron2023llama2} excel in many downstream applications, 
but their limited context window restricts their performance in tasks that require long documents \cite{shaham-etal-2023-zeroscrolls} or a large number of retrieved passages \cite{gao2023alce}.
It is challenging to extend these models to longer context windows 
directly through fine-tuning
due to the scarcity of high-quality instruction data.

To this end, 
we propose \ourschat{}, which uses an auxiliary distillation loss to 
encourage $\menc$ and the cross-attention layers to learn the capabilities of the already instruction-tuned $\mdec$.
This can be especially useful for settings where the fine-tuning data are not open-sourced, which is the case for \llamachat{} \cite{touvron2023llama2}.

\paragraph{Distillation loss.}
We design a distillation objective, where the original $\mdec$ acts as the ``teacher'' and the \ourschat{} model 
acts as the ``student''. 
We use the input context $\concat(\mathcal{C}, X)$ with $m+n=4,096$ tokens, which can fit in the context window of \llamachat, our choice of $\mdec$.
First, we input $\concat(\mathcal{C}, X)$ to $\mdec$ and save the logits of $X$ as the teacher logits.
During training, $\mathcal{C}$ and $X$ are used as inputs to $\menc$ and $\mdec$, respectively, 
and we
minimize the KL divergence between the output logits of $X$ and the corresponding teacher logits as well as the cross-entropy loss. %
We train the model for 10B tokens from our data mixture.
For more details, see \S\ref{app:training_kl}.

\begin{table*}[t]
    \centering
    \small
    \begin{tabular}{lcccccccc}
        \toprule
        & \textbf{ArXiv}           & \textbf{Book}          & \textbf{PG19}     & \textbf{ProofPile} & \textbf{CodeParrot} & \textbf{Throughput} & \textbf{Mem. (GB)}  \\
        \midrule
        \multicolumn{5}{l}{\textbf{Total Tokens $= 4,096$}}                             &                &               &               \\
        \midrule
        \llama     & 2.597          & \textbf{6.282} & 7.614          & 2.409          & \textbf{1.735} & 1.00$\times$           & \textbf{19.2}          \\
        \llamak    & 2.601          & 6.621          & 7.945          & 2.414          & 1.785          & 1.00$\times$          & \textbf{19.2}          \\
        \yarn      & 2.651          & 6.337          & \textbf{7.326} & 2.457          & 1.764          & 1.04$\times$           & \textbf{19.2}          \\
        \ours      & \textbf{2.579} & 6.292          & 7.536          & \textbf{2.396} & 1.763          & \textbf{1.31}$\times$  & 19.8          \\
        \midrule
        \multicolumn{5}{l}{\textbf{Total Tokens $=  8,192$}}                            &                &               &               \\
        \midrule
        \llama            & $>10^3$       & $>10^3$      & $>10^3$ & $>10^3$ & $>10^3$   & -       & - \\
        \llamak    & 2.505          & 6.339          & 7.744          & 2.221          & 1.729          & 1.00$\times$           & 24.9          \\
        \yarn      & 2.561          & 6.077          & \textbf{7.146} & 2.267          & \textbf{1.714}  & 2.52$\times$           & 24.8          \\
        \replug    & 2.589          & 6.149          & 7.554          & 2.307          & 1.728          & 0.17$\times$           & \textbf{18.8} \\
        \streaming & 2.740          & 6.327          & 7.783          & 2.437          & 1.806          & 1.94$\times$           & 20.0          \\
        \ours  & \textbf{2.496} & \textbf{6.049} & 7.372          & \textbf{2.219} & 1.715          & \textbf{3.48}$\times$  & 22.6          \\
        \midrule
        \multicolumn{5}{l}{\textbf{Total Tokens $= 32,768$}}                            &                &               &               \\
        \midrule
        \llamak    & \textbf{2.322} & 6.178          & 7.420          & \textbf{2.158} & 1.664          & 1.00$\times$           & 59.1          \\
        \yarn      & 2.359          & \textbf{5.884} & \textbf{6.809} & 2.193          & \textbf{1.640} & 1.03$\times$           & 58.9          \\
        \streaming & 2.752          & 6.358          & 7.627          & 2.503          & 1.853          & 1.16$\times$           & \textbf{20.0}          \\
        \ours      & 2.421          & 6.015          & 7.204          & 2.218          & 1.702          & \textbf{3.72}$\times$  & 25.6 \\
        \midrule
        \multicolumn{5}{l}{\textbf{Total Tokens $= 131,072$}}                           &                &               &               \\
        \midrule
        \llamak            & $>10^3$        & $>10^3$        & $>10^3$ & $>10^3$ & $>10^3$ & - & - \\
        \yarn              & $>10^3$        & $>10^3$        & $>10^3$ & $>10^3$ & $>10^3$ & - & - \\
        \yarnmore  & 2.359          & 5.270          & 6.306          & 2.242          & \textbf{1.264} & 1.00$\times$           & 235.6         \\
        \streaming & 2.371          & 5.058          & 6.681          & 2.270          & 1.280          & 2.56$\times$           & \textbf{20.0}          \\
        \ours  & \textbf{2.217} & \textbf{4.869} & \textbf{6.305} & \textbf{2.099} & 1.266          & \textbf{9.90}$\times$  & 38.6 \\
       \bottomrule
    \end{tabular}
    \caption{
        Long-context language modeling text perplexity
        on ArXiv and Book from RedPajama, PG19, ProofPile, and CodeParrot. 
        Throughput compares the speed (number of tokens/second) of each model with that of \llama{}.
        All experiments are conducted on one A100 80GB GPU, except for \llamak{} and {\sc{YaRN}} with 128K tokens, which require model parallelism and are conducted on four A100 GPUs.
    }
    \vspace{-2pt}
    \label{tab:eval_lm_ab_32k}
\end{table*}

\section{Long-context Language Modeling}

We start by evaluating \ours{} on long-context language modeling benchmarks 
to assess basic LM abilities, 
and compare with existing approaches in terms of perplexity, memory, and throughput.

\begin{table*}[t]
    \centering
    \small
    \begin{tabular}{lcccccccc}
        \toprule
        & \textbf{ArXiv} & \textbf{Book} & \textbf{C4-RP} & \textbf{CC} & \textbf{Github} & \textbf{StackEx} & \textbf{Wiki} & \textbf{Avg.}\\
        \midrule
        \multicolumn{7}{l}{$k=0$ ($T = 2,048$)} \\
        \midrule
        \llama              & \textbf{3.541}  & \textbf{6.524}  & \textbf{6.916}  & \textbf{5.564}  & \textbf{1.865} & \textbf{4.043} & \textbf{4.816} & \textbf{4.753} \\
        \llamak             & 3.561  & 6.892  & 7.798  & 5.931  & 1.932 & 4.262 & 4.958 & 5.048 \\
        \yarn               & 3.633  & 6.631  & 7.164  & 5.701  & 1.930 & 4.164 & 4.837 & 4.866 \\
        \midrule
        \multicolumn{8}{l}{$k = 8$ ($T = 4,096$)}    &       \\
        \midrule
        \llama              & 3.602  & 6.581  & 6.963  & 5.348  & 1.829 & 4.044 & 4.815 & 4.740 \\
        \llamak             & 3.642  & 6.985  & 7.767  & 5.645  & 1.893 & 4.270 & 4.988 & 5.027 \\
        \yarn               & 3.752  & 6.718  & 7.218  & 5.466  & 1.894 & 4.178 & 4.847 & 4.868 \\
        \replug             & 3.535  & 6.494  & 6.895  & 5.395  & 1.833 & 4.029 & 4.798 & 4.711 \\
        \ours               & \textbf{3.486}  & \textbf{6.481}  & \textbf{6.884}  & \textbf{5.319}  & \textbf{1.793} & \textbf{3.709} & \textbf{4.302} & \textbf{4.568} \\
        \midrule
        \multicolumn{8}{l}{$k = 20$ ($T = 7,168$)}   &       \\
        \midrule
        \replug             & 3.531  & 6.490  & 6.894  & 5.386  & 1.830 & 4.028 & 4.795 & 4.708 \\
        \ours               & \textbf{3.475}  & \textbf{6.463}  & \textbf{6.875}  & \textbf{5.266}  & \textbf{1.782} & \textbf{3.703} & \textbf{4.296} & \textbf{4.551} \\
        \midrule
        \multicolumn{8}{l}{$k = 50$ ($T = 14,848$)}  &       \\
        \midrule
        \replug             & 3.530  & 6.491  & 6.899  & 5.392  & 1.830 & 4.028 & 4.794 & 4.709 \\
        \ours               & \textbf{3.467}  & \textbf{6.457}  & \textbf{6.881}  & \textbf{5.273}  & \textbf{1.777} & \textbf{3.701} & \textbf{4.292} & \textbf{4.550} \\
        \bottomrule
    \end{tabular}
    \caption{
        Retrieval-augmented language modeling. We report test perplexity on RedPajama across all domains.
        We calculate perplexity on the last 1792 tokens of the decoder input (to exclude the query tokens). 
        $k$ is the number of retrieved contexts used, and $T$ is the total number of tokens.
        For \llama{}, \yarn{}, and \llamak, we concatenate the contexts and prepend to the input.
        Avg. is the macro average across all domains. 
    }
    \vspace{-2pt}
    \label{tab:eval_lm_rp}
\end{table*}

\paragraph{Datasets.}
We evaluate on ArXiv and Books from our RedPajama test split, as well as  
three long-context datasets:
PG19 \cite{Rae2020Compressive}, 
Proof-Pile \cite{azerbayev2023proofpile}, 
and CodeParrot \cite{wolf2023codeparrot}. 
We filter all documents to have at least $32,768$ tokens,
and sample $5,000$ sequences 
for each dataset.
We calculate the perplexity on the last $256$ tokens of each sequence. %
Following \citet{peng2024yarn}, 
for the experiments with 128K tokens, we filter documents to have at least $131,072$ tokens and evaluate on only $10$ sequences.\footnote{This is due to computational cost and scarcity of long documents. All $10$ sequences are from different documents.}

\paragraph{Models.}
We focus on 7B parameters models.
Our baseline includes
\llama{} and
its long-sequence fine-tuned versions: 
\llamak~\cite{together2023llama32k},
\yarn{}, and \yarnmore{}~\cite{peng2024yarn}.
\llamak{} was trained on RP with upsampled ArXiv and Books domains, and a mixture of other data 
, and {\sc{YaRN}} was trained on PG19.
While \yarn{} and \yarnmore{} were fine-tuned for only 1.7B and 3.4B tokens, respectively, fine-tuning all parameters requires much more memory relative to \ours{}.
We also evaluate on training-free long-context methods:
\streaming{}~\cite{xiao2024efficient}
and \replug{}~\cite{shi2023large} with \llama{}-7B.
Although \replug{} 
was originally evaluated in a retrieval-augmented setting, we found that it also works well in long-context modeling, 
when viewing the long context as retrieved context.
See \S\ref{app:baseline} for more implementation details.

For \ours, we put 2K tokens in the decoder when $T=4,096$, and 4K tokens in the decoder in other settings. 
Additional tokens are split into chunks of $256$ tokens and fed into the encoder. 
 
\begin{figure*}[t]
    \centering
    \includegraphics[width=0.85\linewidth]{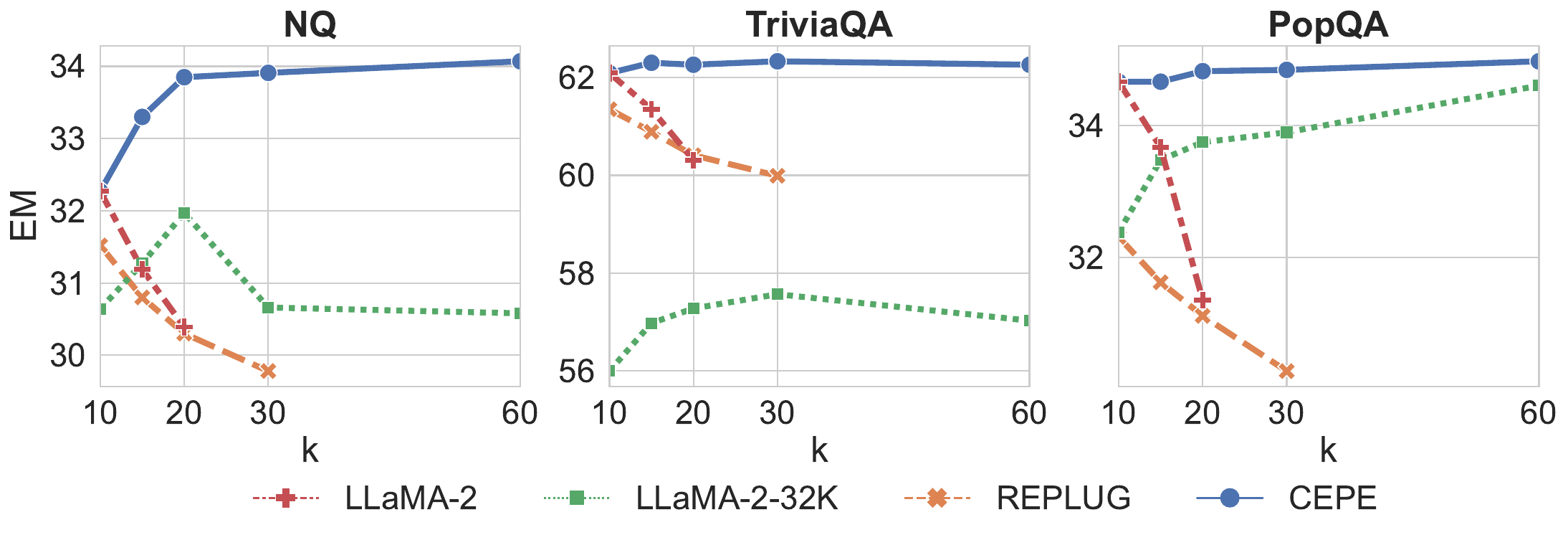}
    \vspace*{-0.3cm}
    \caption{
        Open-domain QA results. 
        We report the exact match (EM) scores.
        \llama{} is limited to 20 passages, and \replug{} is limited to 30 passages due to memory constraints.
        For the complete results, refer to Table \ref{tab:test_qa}.
    }
    \label{fig:test_qa}
    \vspace*{-0.3cm}
\end{figure*}

\paragraph{Results.}
We show the results in Table \ref{tab:eval_lm_ab_32k}.
Compared to the two fully fine-tuned models, 
\llamak{} and \yarn{},
\ours{} achieves either lower or comparable perplexity across all datasets with lower memory usage and higher throughput despite being trained only 8K sequences.
Furthermore, \ours{} continues to improve on perplexity while maintaining low memory use at 128K tokens, well beyond its training lengths (8K); 
on the other hand, \llamak{} and \yarn{} 
cannot generalize beyond its training length and the memory cost increases significantly.
At 128K tokens, we outperform or achieve comparable perplexity with all applicable baselines on all domains.
\revision{Although \ours{} observes higher perplexity than \yarn{} on PG19 and Book at 32K, we note that \yarn{} was fine-tuned on PG19, and thus has a significant domain advantage.}

We also outperform \replug{} across all domains while achieving much higher throughput---\replug{} encodes additional context in chunks but each chunk requires a forward pass of the main input, incurring slow speed, due to which we omit \replug{} at longer length. 
\streaming{} maintains a low memory usage and a reasonable throughput, but the perplexity does not always decrease as the sequence length increases, likely due to the model only using a limited number of cached key-value pairs. %
Compared to \streaming{},
\ours{} can leverage all input tokens and 
achieve better perplexity with better throughput.

\section{Retrieval-augmented Applications}

Retrieval-augmented settings naturally benefit from long-context LMs, as
models can leverage the additional context to include more retrieval results.
Thus, we test if \ours{} trained on long-context data can transfer to retrieval-augmented settings.

\subsection{Retrieval-augmented Language Modeling}

\paragraph{Datasets.}
We use the test and retrieval split of RedPajama described in \S\ref{sec:methods:data} for retrieval-augmented LM evaluation. 
Each sequence contains $2,048$ tokens, and the first $256$ tokens are used as the query to retrieve passages from the retrieval split.
The retrieval corpus contains 200M documents of $256$ tokens each, and we use Contriever \cite{izacard2022unsupervised} to retrieve $k$ passages for each sequence.

\paragraph{Models.}
We evaluate full-context baselines, 
\llama{}, \llamak{}, and \yarn{}, %
by simply prepending the retrieved passages to the input sequence.
We also evaluate \replug{},
which runs one forward pass for each retrieved passage
and aggregates the results.
\ours{} uses $2,048$ tokens in the decoder and
retrieved passages are fed through the
encoder in parallel.

\paragraph{Results.}
The results are shown in Table \ref{tab:eval_lm_rp}.
\ours{} can effectively improve perplexity by using the retrieved contexts, outperforming \replug{}.
Notably, \ours{}
continues to improve perplexity even with $k=50$ (trained with $k=16$). 
\ours{} transfers well to the retrieval-augmented setting
whereas the full-context decoder models degrade. %

\subsection{Open-domain Question Answering}

Given a question and a large corpus of documents, 
open-domain question answering (QA)
requires
the model to retrieve relevant passages and generate the answer.
A model that can leverage a large number of retrieved passages without being distracted by irrelevant ones is desirable for this task.

\begin{table*}[t]
    \centering
    \small
    \resizebox{0.98\linewidth}{!}{
        \setlength{\tabcolsep}{4pt} %
        \begin{tabular}{llccccccccccc}
            \toprule
            & $k$ & SST2 & MR & AGNews & SST5 & TREC & TREC-F & DBPedia & NLU-S & NLU-I & BANKING & CLINIC \\
            \midrule
            \multirow{1}{*}{\llama} & 2  & 89.1 & 96.7 & 72.7 & 3.9 & \textbf{48.0} & 16.7 & \textbf{94.0} & 42.3 & 22.3 & 38.4 & 59.1 \\
            \multirow{2}{*}{+ \ours}& 2 + 18 & 90.7 & \textbf{98.4} & 71.9 & \textbf{46.7} & 47.1 & 22.8 & \textbf{94.0}& \textbf{48.9} & 30.4 & 42.5 & 62.4 \\
                                    & 2 + 38 & \textbf{92.9}& 98.0 & \textbf{73.2} & 45.5 & 47.5 & \textbf{25.1} & 93.3 & 48.8 & \textbf{31.6} & \textbf{46.0} & \textbf{62.8} \\
            \midrule
            \llama$^\dagger$ & 40 & 94.3 & 98.7 & 74.7 & 52.3 & 87.7 & 54.8 & 95.1 & 76.7 & 62.1 & 50.4 & 72.0 \\
            \bottomrule
        \end{tabular}
    }
    \caption{
        ICL results averaged across 3 seeds.
        $k$ is the number of demonstrations.
        All models use 2 demonstrations in the decoder, and we add $+ k'$ 
        demonstrations to the encoder for \ours.
        $^\dagger$ denotes the oracle setting with $k=40$ demonstration in the decoder. 
    }
    \vspace{-2pt}
    \label{tab:eval_icl_dec2}
\end{table*}

\begin{table*}[ht]
    \centering
    \small
        \begin{tabular}{llcccccc}
            \toprule
             & & \multicolumn{3}{c}{\textbf{Question Answering}} & \multicolumn{3}{c}{\textbf{Summarization}} \\ 
            \cmidrule(lr){3-5} \cmidrule(lr){6-8}
            & Total tokens   & NQA & Qspr & QALT & GvRp & SSFD  & QMSum \\
            \midrule
            \llamachat       & 2K & 17.1 & 14.6 & 28.6 & 16.0 & 16.4 & 19.3 \\
            \midrule
            \multirow{3}{*}{+ \ourschat{}}      & 2K + 2K & 19.5 & \textbf{20.5} & \textbf{30.2} & \textbf{16.5} & 16.4 & \textbf{19.6} \\
                                                & 2K + 30K & 21.6 & 19.9 & 29.6 & 15.8 & \textbf{16.7} & 19.5 \\
                                                & 2K + All & \textbf{21.9} & - & - & - & - & - \\
            \midrule
            \llamachat                          & 4K & 18.6 & 16.1 & 30.0 & 17.7 & 17.1 & 19.7 \\
            \llamakchat                         & 32K & 12.2 & 18.1 & 41.6 & 19.9 & 10.0 & 10.3 \\
            \bottomrule
        \end{tabular}
    \caption{
        \zs{} validation results. 
        The total number of tokens includes both the input and generated tokens.
        For NarrativeQA (NQA) and Qspr (QASPER), we report the F1 scores. 
        For QALT (QuALITY), we report accuracy.
        For GovReport (GvRp), SummScreenFD (SSFD), and QMSum, we report the ROUGE-L scores.
        \ourschat{} uses 2K tokens in the decoder, and additional tokens are inputted through the encoder.
        \revision{Only NQA has a substantial number of test examples with more than 32K tokens, so we exclude other dataset on 2K + All setting; see Table~\ref{tab:stats_zs}.}
    }
    \vspace{-0.3em}
    \label{tab:eval_zs}
\end{table*}

\paragraph{Datasets.}
We adopt three open-domain QA datasets: Natural Questions \cite[NQ;][]{kwiatkowski2019natural,lee2019latent}, TriviaQA \cite{joshi2017triviaqa}, and PopQA \cite{mallen-etal-2023-trust}.
For each question, we use Contriever to retrieve $k$ passages from Wikipedia.\footnote{Snapshot from 2018-12-20, and each passage is 100 words \cite{karpukhin-etal-2020-dense}.}

\paragraph{Models.}
We compare \ours{} with \llama{}, \llamak{}, and \replug{}. %
For each model, we use two in-context demonstrations. 
For \ours{}, we use $10$ passages in the decoder, and all other passages are encoded separately by the encoder.
Refer to \S\ref{app:odqa} for more details.

\paragraph{Results.}
The results are shown in Figure \ref{fig:test_qa} 
\ours{} consistently outperforms 
all models 
across all datasets and $k$.
Notably, \ours{} outperforms \llamak{} on NQ and TriviaQA by over 3 and 4 EM points, respectively.
Furthermore, \ours{} does not degrade in performance as the $k$ increases, while other models often perform worse at larger $k$,
as 
they are sensitive to the large amount of redundant or irrelevant passages \cite{Liu2023LostIT}.

\section{In-Context Learning}

In-context learning~\citep[ICL;][]{brown2020language}
is one of the most important emerging qualities of LLMs. 
In this experiment, we examine whether \ours{} can effectively utilize demonstrations from the encoder context to improve performance. %
Specifically, we use 
a range of classification tasks 
that contains a large number (up to 150) of categories \cite{ratner-etal-2023-parallel}, 
where 
the model can benefit from additional demonstrations.
Following \citet{ratner-etal-2023-parallel}, we use a test set size of 250 examples for each dataset.

\paragraph{Models.}
Our baseline is \llama{} with 2 demonstrations. %
For \ours{}, we add additional demonstrations in the encoder. 
We also compare with an ``oracle'', where the \llama{} decoder takes 40 demonstrations.
Note that the oracle is significantly more expensive. 
More details are in \S\ref{app:icl}.

\paragraph{Results.}
The results are shown in Table \ref{tab:eval_icl_dec2}.
    We first observe that compared to the 
    decoder-only baseline,
    \ours{} can effectively use the 
    additional demonstrations from the encoder context;
    the performance further increases or remains consistent with more demonstrations in the encoder. 
However, there is still a large gap to the 
40-demonstration oracle. 
Our hypothesis is that 
in-context learning requires both query-demonstration interactions and 
demonstration-demonstration interactions, which 
\ours{} cannot provide.
Regardless,
\ours{} can be always applied on top of the decoder-only model 
    to add additional demonstrations, with little extra computational and memory cost.

\section{Instruction-tuned Models for Long Text Understanding}
In this section, we show that applying \ourschat{} on \llamachat{} produces instruction-following models that can leverage long inputs.

\paragraph{Datasets.}
\zs{}~\cite{shaham-etal-2023-zeroscrolls}
is a collection of zero-shot long-context understanding tasks that require instruction-following abilities.
Specifically, we test on NarrativeQA, QASPER, QuALITY, GovReport, SummScreenFD, and QMSum, which all have large validation sets made available.
We follow the formats and instructions of \citet{shaham-etal-2023-zeroscrolls} for each dataset, except the long text is placed before the instructions.

\paragraph{Models.}
For \ourschat{}, we use $2,048$ tokens in the decoder and put the remaining tokens to the encoder as chunks of 256 tokens.
We compare with
\llamachat{} and 
\llamakchat{},\footnote{\url{https://huggingface.co/togethercomputer/Llama-2-7B-32K-Instruct}} which was fine-tuned on multi-round conversational data as well as long-context summarization and QA data.
We allow the model to generate $1,024$ tokens for the summarization tasks and $50$ tokens for the question answering tasks.

\paragraph{Results.}
Table \ref{tab:eval_zs} shows that 
\ourschat{} improves upon \llamachat{} with 2K tokens across all tasks.
The performance of \ourschat{} improves or remains consistent as we scale up the number of tokens in the context window. %
Notably, 
\revision{on NQA---the only dataset with a significant number of samples longer than 32K tokens (see Table \ref{tab:stats_zs})---\ourschat{} improves upon \llamachat{} by 3 points in F1 scores.} 
Furthermore, \ourschat{} outperforms \llamakchat{} on 4 out of the 6 tasks, despite being trained on unlabeled data. 
We provide qualitative examples and analysis in \S\ref{app:zs}.

\section{Ablation Studies}

We conduct comprehensive ablations to show
the effectiveness of our training data mixture,
pre-training and fine-tuning the encoder, and the warmup training stage.
We also ablate to verify the effectiveness of the KL divergence loss in \ourschat{}.
\revision{Lastly, we evaluate on Needle in the Haystack \cite{gkamradt_llmtest_needleinahaystack_2024}, and analyze the results in \S\ref{app:needle}.}

\subsection{Training Settings}
\label{sec:ablations:training}

\begin{table}[th]
    \centering
    \resizebox{0.98\linewidth}{!}{
        \begin{tabular}{lcccc}
            \toprule
            & \multicolumn{2}{c}{\textbf{Long-context}} & \multicolumn{2}{c}{\textbf{Retrieval-}} \\ 
            & \multicolumn{2}{c}{\textbf{(total Tokens)}} & \multicolumn{2}{c}{\textbf{augmented ($k$)}} \\ 
            \cmidrule(lr){2-3} \cmidrule(lr){4-5}
                           & \textbf{8K} & \textbf{32K} & \textbf{8}      & \textbf{50}         \\
            \midrule
            \ours             & 3.97 & 3.91 & 4.57 & 4.55 \\
            \midrule
            w/ \retdoc        & 4.01 & 3.99 &{4.53} & {4.50} \\
            w/ \rpcat{} only     & 4.01 & 3.96 & 4.56 & 4.54 \\
            w/ \rpfilter{} only       & {3.96} & {3.89} & 4.75 & 4.72 \\
            \midrule
            w/ Frozen encoder & 4.01 & 3.99 & 4.62 & 4.61 \\
            w/ Random encoder & 4.03 & 4.02 & 4.60 & 4.60 \\
            w/o Warmup   & 4.03 & 4.02 & 4.61 & 4.61 \\
            \bottomrule
       \end{tabular}
    }
    \caption{
        Test perplexity in long-context and retrieval-augmented language modeling, averaged over all datasets.
        Full results are in \Cref{tab:ab_prevdoc_full} and \Cref{tab:ab_retdoc_full}.
    }
    \vspace{-1em}
    \label{tab:ab_retdoc}
\end{table}

\paragraph{Training with retrieved documents.}
Even though \ours{} was trained with long documents, it achieves strong performance on retrieval-augmented applications.
We also test a different data strategy: training \ours{} with retrieved documents using the training settings from \S\ref{sec:methods:training}.

As shown in Table \ref{tab:ab_retdoc}, we find that training on retrieved documents (\retdoc) results in slightly stronger performances in the retrieval-augmentation setting, but performs worse on long documents.
Augmenting a pre-training corpus with retrieval contexts is extremely computationally and storage expensive.
\ours{}'s simple long-document data strategy
achieves a good balance 
between efficient training
and strong performance on 
both long-context and retrieval-augmented applications.

\paragraph{Choices of unlabeled data.}
In Table \ref{tab:ab_retdoc}, we show the results of training \ours{} with only the filtered documents from the ArXiv and Books domains (\textbf{\rpfilter{}}), and only the concatenated RP documents (\textbf{\rpcat{}}).
We find that training on
\rpfilter{} is more beneficial for the long-document setting
and training on \rpcat{} is better for the retrieval setting,
but using a mixture of both
leads to a more balanced and generalizable model.
Our findings corroborate with the recent work on long-context data engineering \cite{fu2024data}. 
We also ablate different training strategies for the encoder, and find both the warmup stage and fine-tuning are crucial for strong performance.
More details on the training ablations are in \S\ref{app:ablations:training}.

\subsection{KL Divergence}

The key component of \ourschat{} is the KL Divergence loss.
To understand the importance of this auxiliary loss, we explore the performance of \ourschat{} when trained without the KL Divergence loss as well as with difference coefficients for each loss. 
Results are shown in Table \ref{tab:ab_zs}, and we find that
the KL Divergence loss is crucial for summarization tasks and QALT.
More details are in \S\ref{app:ablations:kl}.

\section{Related Work}
\label{sec:related_works}

\paragraph{Long-context language models.}
Many recent works on long-context LMs aim to solve the problem of positional embedding extrapolation in transformers \cite{peng2024yarn,chen2023extending}.
Others fine-tune LMs on longer sequences \cite{xiong2023effective,chen2024longlora,together2023llama32k} or compress the context into shorter forms \cite{yoshida2020adding,choromanski2021rethinking,chevalier-etal-2023-adapting}.
Notably, 
several recent papers propose to extend the context window of LMs by modifying the attention mechanism:
\citet{xiao2024efficient} 
discover the use of ``sink tokens'' in sliding windows, and 
\revision{\citet{bertsch2023unlimiformer,xiao2024infllm}
retrieve relevant tokens from a cache}
instead of attending to all tokens.
This results in memory-efficient long-context LMs, but they have diminishing returns with longer contexts, as the same positional embedding may be seen multiple times and they can not fully utilize all tokens.
The key advantage of \ours{} is that it does not degrade for inputs longer than the training length while achieving great efficiency compared to full fine-tuning approaches.
\revision{Techniques have been designed for specific applications, such as for in-context learning \cite{ratner-etal-2023-parallel,hao2022structured}, but we focus on general long-context language modeling.}

Novel architectures and pre-training methods, 
such as S4 \cite{gu2022efficiently}, RPT \cite{rubin2023longrange}, YOCO \cite{sun2024cache}, and Mamba \cite{gu2023mamba}, also extend the context window at greater efficiency.
However, pre-training is extremely expensive at scale and these methods cannot leverage existing powerful pre-trained LLMs.
It is also unclear if state-space models can achieve comparable performance with transformers \cite{jelassi2024repeat}.

\paragraph{Retrieval-augmented language models.}
Augmenting LMs with retrieval has been useful in many applications, such as open-domain question answering.
Recently, combining LMs with retrieval systems for more generalized purposes has been explored: 
\citet{guu2020realm,borgeaud2022retro,izacard2022atlas,min-etal-2023-nonparametric} pre-train LMs with retrieval, 
and \citet{shi2023replug,lin2023radit} use logits interpolation from separate forward passes to incorporate retrieval information.

Our architecture is similar to Atlas \cite{izacard2022atlas}, \retro, and \retro-Fitting~\cite{borgeaud2022retro}
, but they use retrieval-augmented data for pre-training, which can be expensive to acquire at a large scale.
\ours{} only requires fine-tuning on long document data, which are much more efficient to obtain; 
\ours{} is also applicable to any decoder-only LM, allowing us to extend context windows 
for pre-existing strong models.
Finally, these works do not consider long-context language modeling settings in addition to retrieval-augmented settings.

\section{Conclusion}

We propose \ours{} to extend the context window of existing language models.
The key idea is to leverage a small encoder and cross-attention modules to process long inputs and achieve low memory and computational complexity. 
Compared to 
existing methods, 
\ours{} extrapolates to input lengths well beyond the training length,
while remaining efficient and effective. 
Consequently, \ours{} augments pre-trained models to be performant on both long-context and retrieval-augmented applications. 
We also show that \ourschat{} can be applied to instruction-tuned models with additional contexts using an auxiliary loss with only unlabeled data.
We believe that there is still room for improvement through better data to train flexible and robust models.
We hope our work can be a useful and accessible tool for the community to study long-context models in diverse applications.

\section*{Limitations}
One limitation of our work is the focus on \llama-7B.
We hope that future work can investigate the applicability of our framework to a wider variety of LLMs of different sizes. 
Similarly, we only applied \ourschat{} to \llamachat{}-7B, but we look forward to other researchers applying it to other instruction-tuned or fine-tuned models. 

We also acknowledge that certain hyperparameters are not studied in depth due to training costs -- such as the ratio between \rpfilter{} and \rpcat{}, learning rate, and the size of the small encoder model. 
We also fixed Contriever \cite{izacard2022unsupervised} to be the retriever in this work, but studying a greater range of retrievers would be useful.

\section*{Ethics Statement}
LLMs are known to potentially output harmful and/or offensive language,
and the \llama{}-based models we use in this work are no exceptions.
Since these models are trained on internet-size corpora (e.g., RedPajama),
it can be difficult and expensive to filter out such offensive language.

Our models are also fine-tuned on RedPajama, which means they may also generate undesirable language.
Although addressing this issue in the large-scale pre-training corpus is out of the scope of this work, we hope that future work will carefully resolve possible misuse issues in these models.

\section*{Acknowledgements}
We want to acknowledge Sadhika Malladi, Alexander Wettig, Mengzhou Xia, Dan Friedman, Amanda Bertsch, Adithya Bhaskar, Victoria Graf, and other members of the Princeton NLP group for their useful feedback and discussion.
Tianyu Gao is supported by an IBM PhD Fellowship.
This work is also gratefully supported by an NSF CAREER Award (IIS-2239290) and Intel.

\bibliography{anthology,custom}

\begin{thebibliography}{79}
\expandafter\ifx\csname natexlab\endcsname\relax\def\natexlab#1{#1}\fi

\bibitem[{Askell et~al.(2021)Askell, Bai, Chen, Drain, Ganguli, Henighan, Jones, Joseph, Mann, DasSarma, Elhage, Hatfield-Dodds, Hernandez, Kernion, Ndousse, Olsson, Amodei, Brown, Clark, McCandlish, Olah, and Kaplan}]{askell2021general}
Amanda Askell, Yuntao Bai, Anna Chen, Dawn Drain, Deep Ganguli, Tom Henighan, Andy Jones, Nicholas Joseph, Ben Mann, Nova DasSarma, Nelson Elhage, Zac Hatfield-Dodds, Danny Hernandez, Jackson Kernion, Kamal Ndousse, Catherine Olsson, Dario Amodei, Tom Brown, Jack Clark, Sam McCandlish, Chris Olah, and Jared Kaplan. 2021.
\newblock \href {http://arxiv.org/abs/2112.00861} {A general language assistant as a laboratory for alignment}.

\bibitem[{Azerbayev et~al.(2023)Azerbayev, Ayers, and Piotrowski}]{azerbayev2023proofpile}
Zhangir Azerbayev, Edward Ayers, and Bartosz Piotrowski. 2023.
\newblock \href {https://huggingface.co/datasets/hoskinson-center/proof-pile} {Proofpile: A pre-training dataset of mathematical texts}.

\bibitem[{Bai et~al.(2022)Bai, Jones, Ndousse, Askell, Chen, DasSarma, Drain, Fort, Ganguli, Henighan, Joseph, Kadavath, Kernion, Conerly, El-Showk, Elhage, Hatfield-Dodds, Hernandez, Hume, Johnston, Kravec, Lovitt, Nanda, Olsson, Amodei, Brown, Clark, McCandlish, Olah, Mann, and Kaplan}]{bai2022training}
Yuntao Bai, Andy Jones, Kamal Ndousse, Amanda Askell, Anna Chen, Nova DasSarma, Dawn Drain, Stanislav Fort, Deep Ganguli, Tom Henighan, Nicholas Joseph, Saurav Kadavath, Jackson Kernion, Tom Conerly, Sheer El-Showk, Nelson Elhage, Zac Hatfield-Dodds, Danny Hernandez, Tristan Hume, Scott Johnston, Shauna Kravec, Liane Lovitt, Neel Nanda, Catherine Olsson, Dario Amodei, Tom Brown, Jack Clark, Sam McCandlish, Chris Olah, Ben Mann, and Jared Kaplan. 2022.
\newblock \href {http://arxiv.org/abs/2204.05862} {Training a helpful and harmless assistant with reinforcement learning from human feedback}.

\bibitem[{Bertsch et~al.(2023)Bertsch, Alon, Neubig, and Gormley}]{bertsch2023unlimiformer}
Amanda Bertsch, Uri Alon, Graham Neubig, and Matthew~R. Gormley. 2023.
\newblock \href {https://openreview.net/forum?id=lJWUJWLCJo} {Unlimiformer: Long-range transformers with unlimited length input}.
\newblock In \emph{Advances in Neural Information Processing Systems (NeurIPS)}.

\bibitem[{Borgeaud et~al.(2022)Borgeaud, Mensch, Hoffmann, Cai, Rutherford, Millican, Van Den~Driessche, Lespiau, Damoc, Clark, De~Las~Casas, Guy, Menick, Ring, Hennigan, Huang, Maggiore, Jones, Cassirer, Brock, Paganini, Irving, Vinyals, Osindero, Simonyan, Rae, Elsen, and Sifre}]{borgeaud2022retro}
Sebastian Borgeaud, Arthur Mensch, Jordan Hoffmann, Trevor Cai, Eliza Rutherford, Katie Millican, George~Bm Van Den~Driessche, Jean-Baptiste Lespiau, Bogdan Damoc, Aidan Clark, Diego De~Las~Casas, Aurelia Guy, Jacob Menick, Roman Ring, Tom Hennigan, Saffron Huang, Loren Maggiore, Chris Jones, Albin Cassirer, Andy Brock, Michela Paganini, Geoffrey Irving, Oriol Vinyals, Simon Osindero, Karen Simonyan, Jack Rae, Erich Elsen, and Laurent Sifre. 2022.
\newblock \href {https://proceedings.mlr.press/v162/borgeaud22a.html} {Improving language models by retrieving from trillions of tokens}.
\newblock In \emph{International Conference on Machine Learning (ICML)}, volume 162, pages 2206--2240.

\bibitem[{Brown et~al.(2020)Brown, Mann, Ryder, Subbiah, Kaplan, Dhariwal, Neelakantan, Shyam, Sastry, Askell et~al.}]{brown2020language}
Tom~B Brown, Benjamin Mann, Nick Ryder, Melanie Subbiah, Jared Kaplan, Prafulla Dhariwal, Arvind Neelakantan, Pranav Shyam, Girish Sastry, Amanda Askell, et~al. 2020.
\newblock \href {https://papers.nips.cc/paper/2020/hash/1457c0d6bfcb4967418bfb8ac142f64a-Abstract.html} {Language models are few-shot learners}.
\newblock In \emph{Advances in Neural Information Processing Systems (NeurIPS)}.

\bibitem[{Casanueva et~al.(2020)Casanueva, Temčinas, Gerz, Henderson, and Vulić}]{casanueva2020efficient}
Iñigo Casanueva, Tadas Temčinas, Daniela Gerz, Matthew Henderson, and Ivan Vulić. 2020.
\newblock \href {http://arxiv.org/abs/2003.04807} {Efficient intent detection with dual sentence encoders}.

\bibitem[{Chen et~al.(2022)Chen, Chu, Wiseman, and Gimpel}]{chen-etal-2022-summscreen}
Mingda Chen, Zewei Chu, Sam Wiseman, and Kevin Gimpel. 2022.
\newblock \href {https://doi.org/10.18653/v1/2022.acl-long.589} {{S}umm{S}creen: A dataset for abstractive screenplay summarization}.
\newblock In \emph{Proceedings of the 60th Annual Meeting of the Association for Computational Linguistics (Volume 1: Long Papers)}, pages 8602--8615, Dublin, Ireland. Association for Computational Linguistics.

\bibitem[{Chen et~al.(2023)Chen, Wong, Chen, and Tian}]{chen2023extending}
Shouyuan Chen, Sherman Wong, Liangjian Chen, and Yuandong Tian. 2023.
\newblock \href {http://arxiv.org/abs/2306.15595} {Extending context window of large language models via positional interpolation}.

\bibitem[{Chen et~al.(2024)Chen, Qian, Tang, Lai, Liu, Han, and Jia}]{chen2024longlora}
Yukang Chen, Shengju Qian, Haotian Tang, Xin Lai, Zhijian Liu, Song Han, and Jiaya Jia. 2024.
\newblock \href {https://openreview.net/forum?id=6PmJoRfdaK} {Longlo{RA}: Efficient fine-tuning of long-context large language models}.
\newblock In \emph{The Twelfth International Conference on Learning Representations}.

\bibitem[{Chevalier et~al.(2023)Chevalier, Wettig, Ajith, and Chen}]{chevalier-etal-2023-adapting}
Alexis Chevalier, Alexander Wettig, Anirudh Ajith, and Danqi Chen. 2023.
\newblock \href {https://doi.org/10.18653/v1/2023.emnlp-main.232} {Adapting language models to compress contexts}.
\newblock In \emph{Proceedings of the 2023 Conference on Empirical Methods in Natural Language Processing}, pages 3829--3846, Singapore. Association for Computational Linguistics.

\bibitem[{Choromanski et~al.(2021)Choromanski, Likhosherstov, Dohan, Song, Gane, Sarlos, Hawkins, Davis, Mohiuddin, Kaiser, Belanger, Colwell, and Weller}]{choromanski2021rethinking}
Krzysztof~Marcin Choromanski, Valerii Likhosherstov, David Dohan, Xingyou Song, Andreea Gane, Tamas Sarlos, Peter Hawkins, Jared~Quincy Davis, Afroz Mohiuddin, Lukasz Kaiser, David~Benjamin Belanger, Lucy~J Colwell, and Adrian Weller. 2021.
\newblock \href {https://openreview.net/forum?id=Ua6zuk0WRH} {Rethinking attention with performers}.
\newblock In \emph{International Conference on Learning Representations}.

\bibitem[{Dasigi et~al.(2021)Dasigi, Lo, Beltagy, Cohan, Smith, and Gardner}]{dasigi2021qasper}
Pradeep Dasigi, Kyle Lo, Iz~Beltagy, Arman Cohan, Noah~A. Smith, and Matt Gardner. 2021.
\newblock \href {https://doi.org/10.18653/v1/2021.naacl-main.365} {A dataset of information-seeking questions and answers anchored in research papers}.
\newblock In \emph{Proceedings of the 2021 Conference of the North American Chapter of the Association for Computational Linguistics: Human Language Technologies}, pages 4599--4610, Online. Association for Computational Linguistics.

\bibitem[{Devlin et~al.(2019)Devlin, Chang, Lee, and Toutanova}]{devlin2019bert}
Jacob Devlin, Ming-Wei Chang, Kenton Lee, and Kristina Toutanova. 2019.
\newblock \href {https://aclanthology.org/N19-1423/} {{BERT}: Pre-training of deep bidirectional {Transformers} for language understanding}.
\newblock In \emph{North American Chapter of the Association for Computational Linguistics (NAACL)}.

\bibitem[{Fu et~al.(2024)Fu, Panda, Niu, Yue, Hajishirzi, Kim, and Peng}]{fu2024data}
Yao Fu, Rameswar Panda, Xinyao Niu, Xiang Yue, Hannaneh Hajishirzi, Yoon Kim, and Hao Peng. 2024.
\newblock \href {http://arxiv.org/abs/2402.10171} {Data engineering for scaling language models to 128k context}.

\bibitem[{Gao et~al.(2023)Gao, Yen, Yu, and Chen}]{gao2023alce}
Tianyu Gao, Howard Yen, Jiatong Yu, and Danqi Chen. 2023.
\newblock \href {https://doi.org/10.18653/v1/2023.emnlp-main.398} {Enabling large language models to generate text with citations}.
\newblock In \emph{Proceedings of the 2023 Conference on Empirical Methods in Natural Language Processing}, pages 6465--6488, Singapore. Association for Computational Linguistics.

\bibitem[{Gu and Dao(2023)}]{gu2023mamba}
Albert Gu and Tri Dao. 2023.
\newblock \href {http://arxiv.org/abs/2312.00752} {Mamba: Linear-time sequence modeling with selective state spaces}.

\bibitem[{Gu et~al.(2022)Gu, Goel, and Re}]{gu2022efficiently}
Albert Gu, Karan Goel, and Christopher Re. 2022.
\newblock \href {https://openreview.net/forum?id=uYLFoz1vlAC} {Efficiently modeling long sequences with structured state spaces}.
\newblock In \emph{International Conference on Learning Representations}.

\bibitem[{Guu et~al.(2020)Guu, Lee, Tung, Pasupat, and Chang}]{guu2020realm}
Kelvin Guu, Kenton Lee, Zora Tung, Panupong Pasupat, and Ming-Wei Chang. 2020.
\newblock \href {https://arxiv.org/pdf/2002.08909.pdf} {{REALM}: Retrieval-augmented language model pre-training}.
\newblock In \emph{International Conference on Machine Learning (ICML)}.

\bibitem[{Han et~al.(2023)Han, Hao, Dong, Sun, and Wei}]{han2023prototypical}
Zhixiong Han, Yaru Hao, Li~Dong, Yutao Sun, and Furu Wei. 2023.
\newblock \href {https://openreview.net/forum?id=nUsP9lFADUF} {Prototypical calibration for few-shot learning of language models}.
\newblock In \emph{The Eleventh International Conference on Learning Representations}.

\bibitem[{Hao et~al.(2022)Hao, Sun, Dong, Han, Gu, and Wei}]{hao2022structured}
Yaru Hao, Yutao Sun, Li~Dong, Zhixiong Han, Yuxian Gu, and Furu Wei. 2022.
\newblock \href {http://arxiv.org/abs/2212.06713} {Structured prompting: Scaling in-context learning to 1,000 examples}.

\bibitem[{Holtzman et~al.(2020)Holtzman, Buys, Du, Forbes, and Choi}]{Holtzman2020The}
Ari Holtzman, Jan Buys, Li~Du, Maxwell Forbes, and Yejin Choi. 2020.
\newblock \href {https://openreview.net/forum?id=rygGQyrFvH} {The curious case of neural text degeneration}.
\newblock In \emph{International Conference on Learning Representations (ICLR)}.

\bibitem[{Holtzman et~al.(2021)Holtzman, West, Shwartz, Choi, and Zettlemoyer}]{holtzman-etal-2021-surface}
Ari Holtzman, Peter West, Vered Shwartz, Yejin Choi, and Luke Zettlemoyer. 2021.
\newblock \href {https://doi.org/10.18653/v1/2021.emnlp-main.564} {Surface form competition: Why the highest probability answer isn{'}t always right}.
\newblock In \emph{Proceedings of the 2021 Conference on Empirical Methods in Natural Language Processing}, pages 7038--7051, Online and Punta Cana, Dominican Republic. Association for Computational Linguistics.

\bibitem[{Huang et~al.(2021)Huang, Cao, Parulian, Ji, and Wang}]{huang2021govreport}
Luyang Huang, Shuyang Cao, Nikolaus Parulian, Heng Ji, and Lu~Wang. 2021.
\newblock \href {https://doi.org/10.18653/v1/2021.naacl-main.112} {Efficient attentions for long document summarization}.
\newblock In \emph{Proceedings of the 2021 Conference of the North American Chapter of the Association for Computational Linguistics: Human Language Technologies}, pages 1419--1436, Online. Association for Computational Linguistics.

\bibitem[{Ivgi et~al.(2023)Ivgi, Shaham, and Berant}]{ivgi2023sled}
Maor Ivgi, Uri Shaham, and Jonathan Berant. 2023.
\newblock \href {https://doi.org/10.1162/tacl_a_00547} {{Efficient Long-Text Understanding with Short-Text Models}}.
\newblock \emph{Transactions of the Association for Computational Linguistics}, 11:284--299.

\bibitem[{Ivison et~al.(2023)Ivison, Wang, Pyatkin, Lambert, Peters, Dasigi, Jang, Wadden, Smith, Beltagy, and Hajishirzi}]{ivison2023camels}
Hamish Ivison, Yizhong Wang, Valentina Pyatkin, Nathan Lambert, Matthew Peters, Pradeep Dasigi, Joel Jang, David Wadden, Noah~A. Smith, Iz~Beltagy, and Hannaneh Hajishirzi. 2023.
\newblock \href {http://arxiv.org/abs/2311.10702} {Camels in a changing climate: Enhancing lm adaptation with {Tulu} 2}.

\bibitem[{Izacard et~al.(2022{\natexlab{a}})Izacard, Caron, Hosseini, Riedel, Bojanowski, Joulin, and Grave}]{izacard2022unsupervised}
Gautier Izacard, Mathilde Caron, Lucas Hosseini, Sebastian Riedel, Piotr Bojanowski, Armand Joulin, and Edouard Grave. 2022{\natexlab{a}}.
\newblock \href {https://openreview.net/forum?id=jKN1pXi7b0} {Unsupervised dense information retrieval with contrastive learning}.
\newblock \emph{Transactions on Machine Learning Research}.

\bibitem[{Izacard and Grave(2021)}]{izacard2021leveraging}
Gautier Izacard and Edouard Grave. 2021.
\newblock \href {https://doi.org/10.18653/v1/2021.eacl-main.74} {Leveraging passage retrieval with generative models for open domain question answering}.
\newblock In \emph{Proceedings of the 16th Conference of the European Chapter of the Association for Computational Linguistics: Main Volume}, pages 874--880, Online. Association for Computational Linguistics.

\bibitem[{Izacard et~al.(2022{\natexlab{b}})Izacard, Lewis, Lomeli, Hosseini, Petroni, Schick, Dwivedi-Yu, Joulin, Riedel, and Grave}]{izacard2022atlas}
Gautier Izacard, Patrick Lewis, Maria Lomeli, Lucas Hosseini, Fabio Petroni, Timo Schick, Jane Dwivedi-Yu, Armand Joulin, Sebastian Riedel, and Edouard Grave. 2022{\natexlab{b}}.
\newblock \href {https://arxiv.org/pdf/2208.03299.pdf} {Atlas: Few-shot learning with retrieval augmented language models}.
\newblock \emph{arXiv preprint arXiv:2208.03299}.

\bibitem[{Jelassi et~al.(2024)Jelassi, Brandfonbrener, Kakade, and Malach}]{jelassi2024repeat}
Samy Jelassi, David Brandfonbrener, Sham~M. Kakade, and Eran Malach. 2024.
\newblock \href {http://arxiv.org/abs/2402.01032} {Repeat after me: Transformers are better than state space models at copying}.

\bibitem[{Joshi et~al.(2017)Joshi, Choi, Weld, and Zettlemoyer}]{joshi2017triviaqa}
Mandar Joshi, Eunsol Choi, Daniel Weld, and Luke Zettlemoyer. 2017.
\newblock \href {https://doi.org/10.18653/v1/P17-1147} {{T}rivia{QA}: A large scale distantly supervised challenge dataset for reading comprehension}.
\newblock In \emph{Proceedings of the 55th Annual Meeting of the Association for Computational Linguistics (Volume 1: Long Papers)}, pages 1601--1611, Vancouver, Canada. Association for Computational Linguistics.

\bibitem[{Kamradt(2024)}]{gkamradt_llmtest_needleinahaystack_2024}
Garrett Kamradt. 2024.
\newblock \href {https://github.com/gkamradt/LLMTest_NeedleInAHaystack} {Needle in a haystack - pressure testing {LLMs}}.

\bibitem[{Karpukhin et~al.(2020)Karpukhin, Oguz, Min, Lewis, Wu, Edunov, Chen, and Yih}]{karpukhin-etal-2020-dense}
Vladimir Karpukhin, Barlas Oguz, Sewon Min, Patrick Lewis, Ledell Wu, Sergey Edunov, Danqi Chen, and Wen-tau Yih. 2020.
\newblock \href {https://doi.org/10.18653/v1/2020.emnlp-main.550} {Dense passage retrieval for open-domain question answering}.
\newblock In \emph{Proceedings of the 2020 Conference on Empirical Methods in Natural Language Processing (EMNLP)}, pages 6769--6781, Online. Association for Computational Linguistics.

\bibitem[{Ko{\v{c}}isk{\'y} et~al.(2018)Ko{\v{c}}isk{\'y}, Schwarz, Blunsom, Dyer, Hermann, Melis, and Grefenstette}]{kocisky2018narrativeqa}
Tom{\'a}{\v{s}} Ko{\v{c}}isk{\'y}, Jonathan Schwarz, Phil Blunsom, Chris Dyer, Karl~Moritz Hermann, G{\'a}bor Melis, and Edward Grefenstette. 2018.
\newblock \href {https://doi.org/10.1162/tacl_a_00023} {The {N}arrative{QA} reading comprehension challenge}.
\newblock \emph{Transactions of the Association for Computational Linguistics}, 6:317--328.

\bibitem[{Kryscinski et~al.(2022)Kryscinski, Rajani, Agarwal, Xiong, and Radev}]{kryscinski-etal-2022-booksum}
Wojciech Kryscinski, Nazneen Rajani, Divyansh Agarwal, Caiming Xiong, and Dragomir Radev. 2022.
\newblock \href {https://aclanthology.org/2022.findings-emnlp.488} {{BOOKSUM}: A collection of datasets for long-form narrative summarization}.
\newblock In \emph{Findings of the Association for Computational Linguistics: EMNLP 2022}, pages 6536--6558, Abu Dhabi, United Arab Emirates. Association for Computational Linguistics.

\bibitem[{Kwiatkowski et~al.(2019)Kwiatkowski, Palomaki, Redfield, Collins, Parikh, Alberti, Epstein, Polosukhin, Devlin, Lee, Toutanova, Jones, Kelcey, Chang, Dai, Uszkoreit, Le, and Petrov}]{kwiatkowski2019natural}
Tom Kwiatkowski, Jennimaria Palomaki, Olivia Redfield, Michael Collins, Ankur Parikh, Chris Alberti, Danielle Epstein, Illia Polosukhin, Jacob Devlin, Kenton Lee, Kristina Toutanova, Llion Jones, Matthew Kelcey, Ming-Wei Chang, Andrew~M. Dai, Jakob Uszkoreit, Quoc Le, and Slav Petrov. 2019.
\newblock \href {https://doi.org/10.1162/tacl_a_00276} {Natural questions: A benchmark for question answering research}.
\newblock \emph{Transactions of the Association for Computational Linguistics}, 7:452--466.

\bibitem[{Larson et~al.(2019)Larson, Mahendran, Peper, Clarke, Lee, Hill, Kummerfeld, Leach, Laurenzano, Tang, and Mars}]{larson-etal-2019-evaluation}
Stefan Larson, Anish Mahendran, Joseph~J. Peper, Christopher Clarke, Andrew Lee, Parker Hill, Jonathan~K. Kummerfeld, Kevin Leach, Michael~A. Laurenzano, Lingjia Tang, and Jason Mars. 2019.
\newblock \href {https://doi.org/10.18653/v1/D19-1131} {An evaluation dataset for intent classification and out-of-scope prediction}.
\newblock In \emph{Proceedings of the 2019 Conference on Empirical Methods in Natural Language Processing and the 9th International Joint Conference on Natural Language Processing (EMNLP-IJCNLP)}, pages 1311--1316, Hong Kong, China. Association for Computational Linguistics.

\bibitem[{Lee et~al.(2019)Lee, Chang, and Toutanova}]{lee2019latent}
Kenton Lee, Ming-Wei Chang, and Kristina Toutanova. 2019.
\newblock \href {https://doi.org/10.18653/v1/P19-1612} {Latent retrieval for weakly supervised open domain question answering}.
\newblock In \emph{Proceedings of the 57th Annual Meeting of the Association for Computational Linguistics}, pages 6086--6096, Florence, Italy. Association for Computational Linguistics (ACL).

\bibitem[{Lin et~al.(2024)Lin, Chen, Chen, Shi, Lomeli, James, Rodriguez, Kahn, Szilvasy, Lewis, Zettlemoyer, and tau Yih}]{lin2023radit}
Xi~Victoria Lin, Xilun Chen, Mingda Chen, Weijia Shi, Maria Lomeli, Richard James, Pedro Rodriguez, Jacob Kahn, Gergely Szilvasy, Mike Lewis, Luke Zettlemoyer, and Wen tau Yih. 2024.
\newblock \href {https://openreview.net/forum?id=22OTbutug9} {{RA}-{DIT}: Retrieval-augmented dual instruction tuning}.
\newblock In \emph{The Twelfth International Conference on Learning Representations}.

\bibitem[{Liu et~al.(2024)Liu, Lin, Hewitt, Paranjape, Bevilacqua, Petroni, and Liang}]{Liu2023LostIT}
Nelson~F. Liu, Kevin Lin, John Hewitt, Ashwin Paranjape, Michele Bevilacqua, Fabio Petroni, and Percy Liang. 2024.
\newblock \href {https://doi.org/10.1162/tacl_a_00638} {{Lost in the Middle: How Language Models Use Long Contexts}}.
\newblock \emph{Transactions of the Association for Computational Linguistics}, 12:157--173.

\bibitem[{Liu et~al.(2019{\natexlab{a}})Liu, Eshghi, Swietojanski, and Rieser}]{Liu2019BenchmarkingNL}
Xingkun Liu, Arash Eshghi, Pawel Swietojanski, and Verena Rieser. 2019{\natexlab{a}}.
\newblock \href {https://api.semanticscholar.org/CorpusID:76660838} {Benchmarking natural language understanding services for building conversational agents}.
\newblock \emph{ArXiv}, abs/1903.05566.

\bibitem[{Liu et~al.(2019{\natexlab{b}})Liu, Ott, Goyal, Du, Joshi, Chen, Levy, Lewis, Zettlemoyer, and Stoyanov}]{liu2019roberta}
Yinhan Liu, Myle Ott, Naman Goyal, Jingfei Du, Mandar Joshi, Danqi Chen, Omer Levy, Mike Lewis, Luke Zettlemoyer, and Veselin Stoyanov. 2019{\natexlab{b}}.
\newblock \href {https://arxiv.org/abs/1907.11692} {{RoBERTa}: {A} robustly optimized {BERT} pretraining approach}.
\newblock \emph{arXiv preprint arXiv:1907.11692}.

\bibitem[{Loshchilov and Hutter(2019)}]{loshchilov2018decoupled}
Ilya Loshchilov and Frank Hutter. 2019.
\newblock \href {https://openreview.net/forum?id=Bkg6RiCqY7} {Decoupled weight decay regularization}.
\newblock In \emph{International Conference on Learning Representations}.

\bibitem[{Lu et~al.(2022)Lu, Bartolo, Moore, Riedel, and Stenetorp}]{lu-etal-2022-fantastically}
Yao Lu, Max Bartolo, Alastair Moore, Sebastian Riedel, and Pontus Stenetorp. 2022.
\newblock \href {https://doi.org/10.18653/v1/2022.acl-long.556} {Fantastically ordered prompts and where to find them: Overcoming few-shot prompt order sensitivity}.
\newblock In \emph{Proceedings of the 60th Annual Meeting of the Association for Computational Linguistics (Volume 1: Long Papers)}, pages 8086--8098, Dublin, Ireland. Association for Computational Linguistics.

\bibitem[{Mallen et~al.(2023)Mallen, Asai, Zhong, Das, Khashabi, and Hajishirzi}]{mallen-etal-2023-trust}
Alex Mallen, Akari Asai, Victor Zhong, Rajarshi Das, Daniel Khashabi, and Hannaneh Hajishirzi. 2023.
\newblock \href {https://doi.org/10.18653/v1/2023.acl-long.546} {When not to trust language models: Investigating effectiveness of parametric and non-parametric memories}.
\newblock In \emph{Association for Computational Linguistics (ACL)}, pages 9802--9822, Toronto, Canada. Association for Computational Linguistics.

\bibitem[{Min et~al.(2023)Min, Shi, Lewis, Chen, Yih, Hajishirzi, and Zettlemoyer}]{min-etal-2023-nonparametric}
Sewon Min, Weijia Shi, Mike Lewis, Xilun Chen, Wen-tau Yih, Hannaneh Hajishirzi, and Luke Zettlemoyer. 2023.
\newblock \href {https://doi.org/10.18653/v1/2023.findings-acl.132} {Nonparametric masked language modeling}.
\newblock In \emph{Findings of the Association for Computational Linguistics: ACL 2023}, pages 2097--2118, Toronto, Canada. Association for Computational Linguistics.

\bibitem[{Ouyang et~al.(2022)Ouyang, Wu, Jiang, Almeida, Wainwright, Mishkin, Zhang, Agarwal, Slama, Ray et~al.}]{ouyang2022training}
Long Ouyang, Jeffrey Wu, Xu~Jiang, Diogo Almeida, Carroll Wainwright, Pamela Mishkin, Chong Zhang, Sandhini Agarwal, Katarina Slama, Alex Ray, et~al. 2022.
\newblock \href {https://proceedings.neurips.cc/paper_files/paper/2022/file/b1efde53be364a73914f58805a001731-Paper-Conference.pdf} {Training language models to follow instructions with human feedback}.
\newblock \emph{Advances in Neural Information Processing Systems (NeurIPS)}, 35:27730--27744.

\bibitem[{Pang and Lee(2005)}]{pang-lee-2005-seeing}
Bo~Pang and Lillian Lee. 2005.
\newblock \href {https://doi.org/10.3115/1219840.1219855} {Seeing stars: Exploiting class relationships for sentiment categorization with respect to rating scales}.
\newblock In \emph{Proceedings of the 43rd Annual Meeting of the Association for Computational Linguistics ({ACL}{'}05)}, pages 115--124, Ann Arbor, Michigan. Association for Computational Linguistics.

\bibitem[{Pang et~al.(2022)Pang, Parrish, Joshi, Nangia, Phang, Chen, Padmakumar, Ma, Thompson, He, and Bowman}]{pang-etal-2022-quality}
Richard~Yuanzhe Pang, Alicia Parrish, Nitish Joshi, Nikita Nangia, Jason Phang, Angelica Chen, Vishakh Padmakumar, Johnny Ma, Jana Thompson, He~He, and Samuel Bowman. 2022.
\newblock \href {https://doi.org/10.18653/v1/2022.naacl-main.391} {{Q}u{ALITY}: Question answering with long input texts, yes!}
\newblock In \emph{Proceedings of the 2022 Conference of the North American Chapter of the Association for Computational Linguistics: Human Language Technologies}, pages 5336--5358, Seattle, United States. Association for Computational Linguistics.

\bibitem[{Peng et~al.(2024)Peng, Quesnelle, Fan, and Shippole}]{peng2024yarn}
Bowen Peng, Jeffrey Quesnelle, Honglu Fan, and Enrico Shippole. 2024.
\newblock \href {https://openreview.net/forum?id=wHBfxhZu1u} {Ya{RN}: Efficient context window extension of large language models}.
\newblock In \emph{The Twelfth International Conference on Learning Representations}.

\bibitem[{Press et~al.(2022)Press, Smith, and Lewis}]{press2022train}
Ofir Press, Noah Smith, and Mike Lewis. 2022.
\newblock \href {https://openreview.net/forum?id=R8sQPpGCv0} {Train short, test long: Attention with linear biases enables input length extrapolation}.
\newblock In \emph{International Conference on Learning Representations (ICLR)}.

\bibitem[{Rae et~al.(2020)Rae, Potapenko, Jayakumar, Hillier, and Lillicrap}]{Rae2020Compressive}
Jack~W. Rae, Anna Potapenko, Siddhant~M. Jayakumar, Chloe Hillier, and Timothy~P. Lillicrap. 2020.
\newblock \href {https://openreview.net/forum?id=SylKikSYDH} {Compressive transformers for long-range sequence modelling}.
\newblock In \emph{International Conference on Learning Representations}.

\bibitem[{Raffel et~al.(2020)Raffel, Shazeer, Roberts, Lee, Narang, Matena, Zhou, Li, and Liu}]{raffel2020exploring}
Colin Raffel, Noam Shazeer, Adam Roberts, Katherine Lee, Sharan Narang, Michael Matena, Yanqi Zhou, Wei Li, and Peter~J Liu. 2020.
\newblock \href {https://jmlr.org/papers/v21/20-074.html} {Exploring the limits of transfer learning with a unified text-to-text {Transformer}}.
\newblock \emph{The Journal of Machine Learning Research (JMLR)}, 21(140).

\bibitem[{Ratner et~al.(2023)Ratner, Levine, Belinkov, Ram, Magar, Abend, Karpas, Shashua, Leyton-Brown, and Shoham}]{ratner-etal-2023-parallel}
Nir Ratner, Yoav Levine, Yonatan Belinkov, Ori Ram, Inbal Magar, Omri Abend, Ehud Karpas, Amnon Shashua, Kevin Leyton-Brown, and Yoav Shoham. 2023.
\newblock \href {https://doi.org/10.18653/v1/2023.acl-long.352} {Parallel context windows for large language models}.
\newblock In \emph{Proceedings of the 61st Annual Meeting of the Association for Computational Linguistics (Volume 1: Long Papers)}, pages 6383--6402, Toronto, Canada. Association for Computational Linguistics.

\bibitem[{Rubin and Berant(2023)}]{rubin2023longrange}
Ohad Rubin and Jonathan Berant. 2023.
\newblock \href {http://arxiv.org/abs/2306.13421} {Long-range language modeling with self-retrieval}.

\bibitem[{Shaham et~al.(2023)Shaham, Ivgi, Efrat, Berant, and Levy}]{shaham-etal-2023-zeroscrolls}
Uri Shaham, Maor Ivgi, Avia Efrat, Jonathan Berant, and Omer Levy. 2023.
\newblock \href {https://doi.org/10.18653/v1/2023.findings-emnlp.536} {{Z}ero{SCROLLS}: A zero-shot benchmark for long text understanding}.
\newblock In \emph{Findings of the Association for Computational Linguistics: EMNLP 2023}, pages 7977--7989, Singapore. Association for Computational Linguistics.

\bibitem[{Shaham et~al.(2022)Shaham, Segal, Ivgi, Efrat, Yoran, Haviv, Gupta, Xiong, Geva, Berant, and Levy}]{shaham-etal-2022-scrolls}
Uri Shaham, Elad Segal, Maor Ivgi, Avia Efrat, Ori Yoran, Adi Haviv, Ankit Gupta, Wenhan Xiong, Mor Geva, Jonathan Berant, and Omer Levy. 2022.
\newblock \href {https://aclanthology.org/2022.emnlp-main.823} {{SCROLLS}: Standardized {C}ompa{R}ison over long language sequences}.
\newblock In \emph{Proceedings of the 2022 Conference on Empirical Methods in Natural Language Processing}, pages 12007--12021, Abu Dhabi, United Arab Emirates. Association for Computational Linguistics.

\bibitem[{Shi et~al.(2023)Shi, Chen, Misra, Scales, Dohan, Chi, Sch{\"a}rli, and Zhou}]{shi2023large}
Freda Shi, Xinyun Chen, Kanishka Misra, Nathan Scales, David Dohan, Ed~Chi, Nathanael Sch{\"a}rli, and Denny Zhou. 2023.
\newblock \href {https://arxiv.org/pdf/2302.00093.pdf?trk=public_post_comment-text} {Large language models can be easily distracted by irrelevant context}.
\newblock In \emph{International Conference on Machine Learning (ICML)}.

\bibitem[{Shi et~al.(2024)Shi, Min, Yasunaga, Seo, James, Lewis, Zettlemoyer, and tau Yih}]{shi2023replug}
Weijia Shi, Sewon Min, Michihiro Yasunaga, Minjoon Seo, Rich James, Mike Lewis, Luke Zettlemoyer, and Wen tau Yih. 2024.
\newblock \href {https://arxiv.org/abs/2301.12652} {{REPLUG}: Retrieval-augmented black-box language models}.
\newblock In \emph{North American Chapter of the Association for Computational Linguistics (NAACL)}.

\bibitem[{Socher et~al.(2013)Socher, Perelygin, Wu, Chuang, Manning, Ng, and Potts}]{socher2013recursive_sst-2}
Richard Socher, Alex Perelygin, Jean Wu, Jason Chuang, Christopher~D. Manning, Andrew Ng, and Christopher Potts. 2013.
\newblock \href {https://aclanthology.org/D13-1170.pdf} {Recursive deep models for semantic compositionality over a sentiment treebank}.
\newblock In \emph{Empirical Methods in Natural Language Processing (EMNLP)}.

\bibitem[{Su et~al.(2021)Su, Lu, Pan, Murtadha, Wen, and Liu}]{su2021roformer}
Jianlin Su, Yu~Lu, Shengfeng Pan, Ahmed Murtadha, Bo~Wen, and Yunfeng Liu. 2021.
\newblock \href {http://arxiv.org/abs/2104.09864} {Roformer: Enhanced transformer with rotary position embedding}.

\bibitem[{Sun et~al.(2024)Sun, Dong, Zhu, Huang, Wang, Ma, Zhang, Wang, and Wei}]{sun2024cache}
Yutao Sun, Li~Dong, Yi~Zhu, Shaohan Huang, Wenhui Wang, Shuming Ma, Quanlu Zhang, Jianyong Wang, and Furu Wei. 2024.
\newblock \href {http://arxiv.org/abs/2405.05254} {You only cache once: Decoder-decoder architectures for language models}.

\bibitem[{Taori et~al.(2023)Taori, Gulrajani, Zhang, Dubois, Li, Guestrin, Liang, and Hashimoto}]{alpaca}
Rohan Taori, Ishaan Gulrajani, Tianyi Zhang, Yann Dubois, Xuechen Li, Carlos Guestrin, Percy Liang, and Tatsunori~B. Hashimoto. 2023.
\newblock \href {https://github.com/tatsu-lab/stanford_alpaca} {{Stanford Alpaca: An Instruction-following LLaMA model}}.

\bibitem[{Together(2023{\natexlab{a}})}]{together2023llama32k}
Together. 2023{\natexlab{a}}.
\newblock \href {https://huggingface.co/togethercomputer/LLaMA-2-7B-32K} {Llama-2-7b-32k}.

\bibitem[{Together(2023{\natexlab{b}})}]{together2023redpajama}
Together. 2023{\natexlab{b}}.
\newblock \href {https://github.com/togethercomputer/RedPajama-Data} {Redpajama: An open source recipe to reproduce llama training dataset}.

\bibitem[{Touvron et~al.(2023{\natexlab{a}})Touvron, Lavril, Izacard, Martinet, Lachaux, Lacroix, Rozi{\`e}re, Goyal, Hambro, Azhar et~al.}]{touvron2023llama}
Hugo Touvron, Thibaut Lavril, Gautier Izacard, Xavier Martinet, Marie-Anne Lachaux, Timoth{\'e}e Lacroix, Baptiste Rozi{\`e}re, Naman Goyal, Eric Hambro, Faisal Azhar, et~al. 2023{\natexlab{a}}.
\newblock \href {https://arxiv.org/pdf/2302.13971.pdf} {{LLaMA: Open and Efficient Foundation Language Models}}.
\newblock \emph{arXiv preprint arXiv:2302.13971}.

\bibitem[{Touvron et~al.(2023{\natexlab{b}})Touvron, Martin, Stone, Albert, Almahairi, Babaei, Bashlykov, Batra, Bhargava, Bhosale, Bikel, Blecher, Ferrer, Chen, Cucurull, Esiobu, Fernandes, Fu, Fu, Fuller, Gao, Goswami, Goyal, Hartshorn, Hosseini, Hou, Inan, Kardas, Kerkez, Khabsa, Kloumann, Korenev, Koura, Lachaux, Lavril, Lee, Liskovich, Lu, Mao, Martinet, Mihaylov, Mishra, Molybog, Nie, Poulton, Reizenstein, Rungta, Saladi, Schelten, Silva, Smith, Subramanian, Tan, Tang, Taylor, Williams, Kuan, Xu, Yan, Zarov, Zhang, Fan, Kambadur, Narang, Rodriguez, Stojnic, Edunov, and Scialom}]{touvron2023llama2}
Hugo Touvron, Louis Martin, Kevin Stone, Peter Albert, Amjad Almahairi, Yasmine Babaei, Nikolay Bashlykov, Soumya Batra, Prajjwal Bhargava, Shruti Bhosale, Dan Bikel, Lukas Blecher, Cristian~Canton Ferrer, Moya Chen, Guillem Cucurull, David Esiobu, Jude Fernandes, Jeremy Fu, Wenyin Fu, Brian Fuller, Cynthia Gao, Vedanuj Goswami, Naman Goyal, Anthony Hartshorn, Saghar Hosseini, Rui Hou, Hakan Inan, Marcin Kardas, Viktor Kerkez, Madian Khabsa, Isabel Kloumann, Artem Korenev, Punit~Singh Koura, Marie-Anne Lachaux, Thibaut Lavril, Jenya Lee, Diana Liskovich, Yinghai Lu, Yuning Mao, Xavier Martinet, Todor Mihaylov, Pushkar Mishra, Igor Molybog, Yixin Nie, Andrew Poulton, Jeremy Reizenstein, Rashi Rungta, Kalyan Saladi, Alan Schelten, Ruan Silva, Eric~Michael Smith, Ranjan Subramanian, Xiaoqing~Ellen Tan, Binh Tang, Ross Taylor, Adina Williams, Jian~Xiang Kuan, Puxin Xu, Zheng Yan, Iliyan Zarov, Yuchen Zhang, Angela Fan, Melanie Kambadur, Sharan Narang, Aurelien Rodriguez, Robert Stojnic, Sergey Edunov, and Thomas Scialom. 2023{\natexlab{b}}.
\newblock \href {http://arxiv.org/abs/2307.09288} {Llama 2: Open foundation and fine-tuned chat models}.

\bibitem[{Vaswani et~al.(2017)Vaswani, Shazeer, Parmar, Uszkoreit, Jones, Gomez, Kaiser, and Polosukhin}]{vaswani2017attention}
Ashish Vaswani, Noam Shazeer, Niki Parmar, Jakob Uszkoreit, Llion Jones, Aidan~N Gomez, {\L}ukasz Kaiser, and Illia Polosukhin. 2017.
\newblock \href {https://papers.nips.cc/paper/2017/hash/3f5ee243547dee91fbd053c1c4a845aa-Abstract.html} {Attention is all you need}.
\newblock \emph{Advances in Neural Information Processing Systems (NIPS)}, 30.

\bibitem[{Voorhees and Tice(2000)}]{voorhees2000building_trec}
Ellen~M. Voorhees and Dawn~M. Tice. 2000.
\newblock \href {https://doi.org/10.1145/345508.345577} {Building a question answering test collection}.
\newblock In \emph{Proceedings of the 23rd Annual International ACM SIGIR Conference on Research and Development in Information Retrieval}, SIGIR '00, page 200–207, New York, NY, USA. Association for Computing Machinery.

\bibitem[{Wang et~al.(2023)Wang, Kordi, Mishra, Liu, Smith, Khashabi, and Hajishirzi}]{wang-etal-2023-self-instruct}
Yizhong Wang, Yeganeh Kordi, Swaroop Mishra, Alisa Liu, Noah~A. Smith, Daniel Khashabi, and Hannaneh Hajishirzi. 2023.
\newblock \href {https://doi.org/10.18653/v1/2023.acl-long.754} {Self-instruct: Aligning language models with self-generated instructions}.
\newblock In \emph{Proceedings of the 61st Annual Meeting of the Association for Computational Linguistics (Volume 1: Long Papers)}, pages 13484--13508, Toronto, Canada. Association for Computational Linguistics.

\bibitem[{Wolf et~al.(2023)Wolf, Ben~Allal, von Werra, Jia, and Zebaze}]{wolf2023codeparrot}
Thomas Wolf, Loubna Ben~Allal, Leandro von Werra, Li~Jia, and Armel Zebaze. 2023.
\newblock \href {https://huggingface.co/datasets/codeparrot/codeparrot-valid-v2-near-dedup} {A dataset of python files from github}.
\newblock \url{https://github.com/huggingface/blog/blob/main/codeparrot.md?version=codeparrot/codeparrot-valid-v2-near-dedup}.

\bibitem[{Wolf et~al.(2020)Wolf, Debut, Sanh, Chaumond, Delangue, Moi, Cistac, Rault, Louf, Funtowicz, Davison, Shleifer, von Platen, Ma, Jernite, Plu, Xu, Le~Scao, Gugger, Drame, Lhoest, and Rush}]{wolf2020transformers}
Thomas Wolf, Lysandre Debut, Victor Sanh, Julien Chaumond, Clement Delangue, Anthony Moi, Pierric Cistac, Tim Rault, Remi Louf, Morgan Funtowicz, Joe Davison, Sam Shleifer, Patrick von Platen, Clara Ma, Yacine Jernite, Julien Plu, Canwen Xu, Teven Le~Scao, Sylvain Gugger, Mariama Drame, Quentin Lhoest, and Alexander Rush. 2020.
\newblock \href {https://doi.org/10.18653/v1/2020.emnlp-demos.6} {Transformers: State-of-the-art natural language processing}.
\newblock In \emph{Proceedings of the 2020 Conference on Empirical Methods in Natural Language Processing: System Demonstrations}, pages 38--45, Online. Association for Computational Linguistics.

\bibitem[{Xiao et~al.(2024{\natexlab{a}})Xiao, Zhang, Han, Xiao, Lin, Zhang, Liu, Han, and Sun}]{xiao2024infllm}
Chaojun Xiao, Pengle Zhang, Xu~Han, Guangxuan Xiao, Yankai Lin, Zhengyan Zhang, Zhiyuan Liu, Song Han, and Maosong Sun. 2024{\natexlab{a}}.
\newblock \href {https://arxiv.org/pdf/2402.04617} {{InfLLM}: Unveiling the intrinsic capacity of {LLMs} for understanding extremely long sequences with training-free memory}.
\newblock \emph{arXiv preprint arXiv:2402.04617}.

\bibitem[{Xiao et~al.(2024{\natexlab{b}})Xiao, Tian, Chen, Han, and Lewis}]{xiao2024efficient}
Guangxuan Xiao, Yuandong Tian, Beidi Chen, Song Han, and Mike Lewis. 2024{\natexlab{b}}.
\newblock \href {https://openreview.net/forum?id=NG7sS51zVF} {Efficient streaming language models with attention sinks}.
\newblock In \emph{The Twelfth International Conference on Learning Representations}.

\bibitem[{Xiong et~al.(2023)Xiong, Liu, Molybog, Zhang, Bhargava, Hou, Martin, Rungta, Sankararaman, Oguz, Khabsa, Fang, Mehdad, Narang, Malik, Fan, Bhosale, Edunov, Lewis, Wang, and Ma}]{xiong2023effective}
Wenhan Xiong, Jingyu Liu, Igor Molybog, Hejia Zhang, Prajjwal Bhargava, Rui Hou, Louis Martin, Rashi Rungta, Karthik~Abinav Sankararaman, Barlas Oguz, Madian Khabsa, Han Fang, Yashar Mehdad, Sharan Narang, Kshitiz Malik, Angela Fan, Shruti Bhosale, Sergey Edunov, Mike Lewis, Sinong Wang, and Hao Ma. 2023.
\newblock \href {http://arxiv.org/abs/2309.16039} {Effective long-context scaling of foundation models}.

\bibitem[{Yoshida et~al.(2020)Yoshida, Ettinger, and Gimpel}]{yoshida2020adding}
Davis Yoshida, Allyson Ettinger, and Kevin Gimpel. 2020.
\newblock \href {http://arxiv.org/abs/2008.07027} {Adding recurrence to pretrained transformers for improved efficiency and context size}.

\bibitem[{Zhang et~al.(2015)Zhang, Zhao, and LeCun}]{Zhang2015CharacterlevelCN}
Xiang Zhang, Junbo~Jake Zhao, and Yann LeCun. 2015.
\newblock \href {https://api.semanticscholar.org/CorpusID:368182} {Character-level convolutional networks for text classification}.
\newblock In \emph{Neural Information Processing Systems}.

\bibitem[{Zhao et~al.(2021)Zhao, Wallace, Feng, Klein, and Singh}]{pmlr-v139-zhao21c}
Zihao Zhao, Eric Wallace, Shi Feng, Dan Klein, and Sameer Singh. 2021.
\newblock \href {https://proceedings.mlr.press/v139/zhao21c.html} {Calibrate before use: Improving few-shot performance of language models}.
\newblock In \emph{Proceedings of the 38th International Conference on Machine Learning}, volume 139 of \emph{Proceedings of Machine Learning Research}, pages 12697--12706. PMLR.

\bibitem[{Zhong et~al.(2021)Zhong, Yin, Yu, Zaidi, Mutuma, Jha, Awadallah, Celikyilmaz, Liu, Qiu, and Radev}]{zhong2021qmsum}
Ming Zhong, Da~Yin, Tao Yu, Ahmad Zaidi, Mutethia Mutuma, Rahul Jha, Ahmed~Hassan Awadallah, Asli Celikyilmaz, Yang Liu, Xipeng Qiu, and Dragomir Radev. 2021.
\newblock \href {https://doi.org/10.18653/v1/2021.naacl-main.472} {{QMS}um: A new benchmark for query-based multi-domain meeting summarization}.
\newblock In \emph{Proceedings of the 2021 Conference of the North American Chapter of the Association for Computational Linguistics: Human Language Technologies}, pages 5905--5921, Online. Association for Computational Linguistics.

\end{thebibliography}

\appendix

\clearpage

\section{Training Details}

\subsection{Pre-training Encoder}
\label{app:training_encoder}

The encoder follows the configuration of RoBERTa-large \cite{liu2019roberta} -- it has 24 layers, a hidden size of 1024, and 16 attention heads. 
However, we use the architecture of \llama, which means that the vocabulary size is different and the attention module contains an additional output projection.
We refer to \citet{liu2019roberta} and \citet{touvron2023llama} for more details.

We pre-trained the encoder for 100K steps on RP using the masked language modeling objective \cite{devlin2019bert}.
We used a batch size of 2048 sequences, where each sequence consisted of 512 tokens.
The learning rate was set to $10^{-3}$ with a warm-up of the first 4\% of the steps. 
We used eight A6000 GPUs with a gradient accumulation of 16. 
Furthermore, we employed a masking rate of 30\% and disabled the next sentence prediction objective.
We always replace the token with the \texttt{[MASK]} token if it is masked instead of replacing it with a random token or the original token.
Finally, we used the AdamW optimizer \cite{loshchilov2018decoupled} with $\beta_1=0.9$, $\beta_2=0.999$, and $\epsilon=10^{-8}$, as implemented by the HuggingFace Transformers library \cite{wolf2020transformers}.

\subsection{Training \ours}
\label{app:training}

The attention module in \llama{} consists of four projection matrices: key, value, query, and output.
In contrast to original transformers \cite{vaswani2017attention}, the output projection matrix is used as an additional attention output projection.
When we first insert the cross-attention layers into the decoder, we initialize the weights of the key, value, and query projection matrices with the respective weights from the decoder's self-attention layer in the same transformer block.
Furthermore, since the hidden dimension of the encoder is smaller than the hidden dimension of the decoder, $d < D$, we use only copy the first $d$ rows of the key and value projection matrices from the self-attention module to the cross-attention module.
Lastly, the output projection matrix is initialized with all zeros.
While we did not investigate this initialization in detail, the intuition is that we want the model to use the tokens from the encoder using a similar mechanism as the decoder uses for its own tokens.
However, we want the model to learn the output projection from scratch, as it may be too disruptive to have doubled the number of attention modules. 

Then, we employ a warmup initialization method that simply trains the model to copy the input tokens from the encoder to the decoder.
Specifically, we use the same inputs $X = \mathcal{C}$ for both $\menc$ and $\mdec$, and $X$ consists of $n=256$ tokens.
However, for the encoder, we chunk $X$ into $k=4$ sequences of $64$ tokens to construct $\mathcal{C}$.
This step was trained for 4K steps with a batch size of 128 and peak learning rate of $5 \times 10^{-4}$, which totals to 131M tokens.
We noticed that the model quickly learned to copy the input tokens from the encoder to the decoder, and the loss was close to zero after just 1K steps. 
The intuition behind this initialization strategy was to instill a strong inductive bias between the encoder input and decoder outputs. 
From our early experiments, we found that this initialization strategy helped stabilize the later training.

Finally, we train \ours{} for 20K steps with a batch size of 128.
We use eight A100 80 GB GPUs with a per-device batch size of 2 and gradient accumulation of 8, which took approximately 750 GPU hours. 
We also use a peak learning rate of $3 \times 10^{-4}$ with a warm-up of 4\% of the steps and a cosine learning rate schedule.
We use the AdamW optimizer with $\beta_1=0.9$, $\beta_2=0.999$, and $\epsilon=10^{-8}$.

During the standard training, we also use masking to inject noise into the parallel encodings.
Intuitively, additional contexts may not always be in chunks of exactly $256$ tokens in practice; for example, in open-domain question answering, the retrieved passages may vary in length.
Furthermore, we may not always use exactly $k=16$ contexts during inference, so we may want to train \ours{} with instances with different $k$.
To this end, for each encoder context $C_i \in \mathcal{C}$, we apply masking with a probability of $0.3$.
When masking, we mask out the entire context $C_i$ with probability $0.1$. 
With probability $0.9$, we mask out the suffix tokens $x_{|C_i|-t+1},\ldots,x_{|C_i|}$, where $x_1,\ldots,x_{|C_i|}$ are the tokens in $C_i$ and $t \sim U(1, |C_i|)$.
We choose to only mask out suffixes to maintain a natural distribution for encoder inputs. 
While we did not study the masking rate extensively, we found in preliminary experiments that masking did not hurt perplexity while improving performance on downstream tasks.
We leave further explorations on masking in the encoder input for future work.

\subsection{Training \ourschat}
\label{app:training_kl}
We leverage an additional distillation loss for \ourschat. 
For encoder input $\mathcal{C}$ and decoder input $X$, we first calculate the logits $\mdec(\text{concat}(\mathcal{C}, X))$ by running forward passes with the original model parameterized by $\mdec$. 
Due to storage constraints, we only store the top 50 likelihoods and their indices in the vocabulary for each token in $X$, following \cite{askell2021general,bai2022training}.

Then, during training, we define the distillation loss as the KL Divergence between the teacher model's probability distribution and the student model's probability distribution for the previously stored top 50 tokens.
Concretely, our distillation loss is defined as follows:
\begin{align*}
    \mathcal{L}_{\textrm{KL}} &= D_{\textrm{KL}}(\mdec(S) || \mathcal{M}_{\mathrm{CEPE}}(\mathcal{C}, X))
\end{align*}
where $S = \text{concat}(\mathcal{C}, X)$, $\mdec(S)$ is the probability distribution of the top 50 tokens for $X$, and $\mathcal{M}_{\mathrm{CEPE}}(\mathcal{C}, X)$ takes $\mathcal{C}$ as the encoder input and $X$ as the decoder input and outputs the probability distribution of the same 50 tokens on $X$. 

Although \citet{bai2022training} also uses an additional category that represents the sum of all other tokens' probabilities, we found that this may cause the KL Divergence to be undefined when the sum of other probabilities is $0$. 
For our main model, we use a coefficient of $2$ in front of $\mathcal{L}_{\textrm{KL}}$ when adding to the cross-entropy loss to calculate the total loss. 
We experiment with this coefficient in \S\ref{sec:ablations:training}.

\section{Baseline Implementations}
\label{app:baseline}

\paragraph{\replug{}.}
Although \replug{} \cite{shi2023replug} was introduced as a method to augment language models with retrieval, we found that the technique of interpolating logits from separate forward passes can also transfer well to the long context setting.
Among the methods that we compare to, \replug{} uniquely improves performance upon the base model in both the long-context and the retrieval-augmented LM settings.
This gives us an additional point of comparison across the two settings.

Following the original authors, we use Contriever~\cite{izacard2022unsupervised} to calculate the scores for each previous context by using the first 256 tokens following the previous context as the query in the long-context setting.
We did not include the additional memory and inference time costs of calculating the Contriever scores in our evaluation. 

\paragraph{\streaming{}.} We follow the implementation of \streaming{} from the original authors\footnote{\url{https://github.com/mit-han-lab/streaming-llm}} \cite{xiao2024efficient}.
Specifically, we use their best settings, where we enable the positional shifts and cache 4 sink tokens and 2044 recent tokens.

\begin{table*}[t]
    \centering
    \small
    \begin{tabular}{lccccccc}
        \toprule
         & \textbf{Stride} & \textbf{ArXiv}           & \textbf{Book}          & \textbf{PG19}     & \textbf{ProofPile} & \textbf{CodeParrot}  \\
        \midrule
        \multicolumn{3}{l}{\textbf{Total Tokens $=  8192$}} &          &           &            &            &      \\
        \midrule
        \multirow{2}{*}{\streaming} & 1    & 2.823 & 6.381 & 7.817 & 2.522 & 1.848 \\
                            & 2048 & 2.740 & 6.327 & 7.783 & 2.437 & 1.806 \\
        
        \bottomrule
    \end{tabular}
    \caption{
        Performance of \streaming{} with different stride lengths.
    }
    \vspace{-2pt}
    \label{tab:streaming}
\end{table*}

The original code evaluates the model using a stride of 1 token at a time, where the cache is updated after every token, but this is not feasible for our large-scale evaluation.
Therefore, we use a stride of 2048 tokens, and we update the cache after each stride.
We show the difference in performance between the two settings in Table \ref{tab:streaming}, and we found that \streaming{} benefits from using a larger stride.
We leave future exploration in this direction to future work.

\section{Evaluation Settings}

\subsection{Open-domain Question Answering}
\label{app:odqa}

The full results for the open-domain question answering experiments are shown in Table \ref{tab:test_qa}.
\replug{} only uses up to $k=30$ passages due to memory constraints, and \llama{} has a window size of 4096, which limits $k$ to 20.
While \llamak{} can use more than $k=60$ passages with a context size of 32K, we only use up to 60 passages due to the cost of generation.
For each demonstration, we only show the top $1$ retrieved passages instead of the top $k$ passages.

\begin{table}[ht]
    \centering
    \resizebox{0.98\linewidth}{!}{
        \begin{tabular}{lccccc}
            \toprule
            & $k$ & NQ & TQA & PopQA \\
            \midrule
            \multirow{5}{*}{\llama}     & 1  & 28.37 & 56.44 & 27.17 \\
                & 5  & 31.91 & 61.08 & 33.83 \\
                & 10 & 32.27 & 62.09 & 34.67 \\
                & 15 & 31.19 & 61.35 & 33.67 \\
                & 20 & 30.39 & 60.31 & 31.35 \\
            \midrule
            \multirow{5}{*}{\llamak} & 10 & 30.64 & 56.00 & 32.38 \\
                & 15 & 31.27 & 56.98 & 33.48 \\
                & 20 & 31.97 & 57.28 & 33.75 \\
                & 30 & 30.66 & 57.57 & 33.90 \\
                & 60 & 30.58 & 57.03 & 34.61 \\
            \midrule
            \multirow{5}{*}{\replug}      & 5  & 31.27 & 61.21 & 32.40 \\
                & 10 & 31.52 & 61.35 & 32.31 \\
                & 15 & 30.80 & 60.89 & 31.62 \\
                & 20 & 30.30 & 60.41 & 31.11 \\
                & 30 & 29.78 & 59.99 & 30.27 \\
            \midrule
            \multirow{5}{*}{\ours}        & 10 & 32.27 & 62.09 & 34.67 \\
                    & 15 & 33.30 & 62.30 & 34.67 \\
                    & 20 & 33.85 & 62.26 & 34.83 \\
                    & 30 & 33.91 & 62.33 & 34.85 \\
                    & 60 & 34.07 & 62.26 & 34.98 \\
            \bottomrule
        \end{tabular}
    }
    \caption{
        Open-domain QA results.
        We report exact match scores for the Natural Questions(NQ) test set, TriviaQA(TQA) validation set, and PopQA test set. 
        All models use two-shot in-context learning. 
        $k$ is the number of retrieved passages, and \ours{} uses the top 10 passages in the decoder and all passages in the encoder.
    }
    \label{tab:test_qa}
\end{table}

\subsection{In-context Learning}
\label{app:icl}
For our in-context learning experiments, we use the datasets commonly used in previous works \cite{pmlr-v139-zhao21c,lu-etal-2022-fantastically,han2023prototypical,ratner-etal-2023-parallel}:
SST-2 \cite{socher2013recursive_sst-2},
MovieReview \cite[MR][]{pang-lee-2005-seeing},
AGNews \cite{Zhang2015CharacterlevelCN},
SST-5 \cite{socher2013recursive_sst-2},
TREC \cite{voorhees2000building_trec},
DBPedia \cite{Zhang2015CharacterlevelCN},
NLU \cite{Liu2019BenchmarkingNL},
BANKING77 \cite{casanueva2020efficient},
CLINIC150 \cite{larson-etal-2019-evaluation}.
We follow the prompts used in \citet{ratner-etal-2023-parallel} for all datasets.
During evaluation, we first calculate the log-likelihood of each option and select the option with the highest likelihood. 
We sample the in-context learning demonstrations from the training set such that each label has an equal number of demonstrations (except for possible remainders). 

Furthermore, we first calculate the accuracy for each dataset using four different metrics: likelihood, likelihood normalized for length, calibrated likelihood, and calibrated likelihood normalized for length.
We calibrate using Domain Conditional PMI \cite{holtzman-etal-2021-surface}, but use the empty string as the domain string for all datasets for simplicity. 
We then choose the metrics that yield the highest score for the \llama{} model in the two-shot setting and apply the same metrics to all other models.
The metrics used for each dataset are shown in Table \ref{tab:app_pcw_dataset}.
In this work, we did not investigate how to best calibrate \ours{} in ICL settings.
We leave these explorations for future work. 

\begin{table}[th]
    \centering
        \small
        \begin{tabular}{lcc}
            \toprule
            & Normalized & Calibrated \\
            \midrule
            SST2    & No         & Yes        \\
            MR      & No         & No         \\
            AGNews  & No         & No         \\
            SST5    & No         & Yes        \\
            TREC    & No         & No         \\
            TREC-F  & No         & No         \\
            DBPedia & Yes        & Yes        \\
            NLU-S   & Yes        & Yes        \\
            NLU-I   & No         & No         \\
            BANKING & No         & No         \\
            CLINIC  & No         & No        \\
            \bottomrule
       \end{tabular}
    \caption{
        Metrics used for each dataset. 
        For normalization, we divide the log-likelihood by the length of the prompt.
        For calibration, we use Domain Conditional PMI \cite{holtzman-etal-2021-surface} with the empty string as the domain string for all datasets for simplicity.
    }
    \label{tab:app_pcw_dataset}
\end{table}

\subsection{\zs}
\label{app:zs}
We use a subset of the \zs{} \cite{shaham-etal-2023-zeroscrolls} with large validation sets: NarrativeQA \cite{kocisky2018narrativeqa}, Qasper \cite{dasigi2021qasper}, QuALITY \cite{pang-etal-2022-quality}, GovReport \cite{huang2021govreport}, SummScreenFD \cite{chen-etal-2022-summscreen}, and QMSum \cite{zhong2021qmsum}.
Specifically, we use the validation sets of these datasets made available by {\sc{Scrolls}} \cite{shaham-etal-2022-scrolls}.

However, we follow the same evaluation setup as \zs, where models are evaluated in the zero-shot setting.
We also use the same evaluation metrics as \zs{} for each dataset.
For the question answering datasets (NarrativeQA, Qasper, and QuALITY), we allow the model to generate up to 50 tokens, and we use greedy decoding.

For the summarization datasets (GovReport, SummScreenFD, and QMSum), we allow the model to generate up to $1,024$ tokens, following the original authors.
For SummScreenFD and QMSum, we use greedy decoding, and for GovReport we use nucleus sampling \cite{Holtzman2020The} with a temperature of $1.0$ and top-$p$ of $0.95$ and a minimum generation length of 10 tokens.

This is because GovReport has a much longer gold summary than the other datasets, and sampling methods are typically used in long-generation settings.
Furthermore, greedy decoding may degenerate. 
The minimum generation length helps prevent trivial outputs, such as empty strings.
To account for the randomness in the sampling method, we averaged GovReport performance over 3 seeded runs, and we found that the standard deviation is less than $0.20$ ROUGE-L scores in all settings. 

Furthermore, we hypothesize that \llamakchat{} may overfit to specific domains such as GovReport due to being trained on BookSum, a summarization dataset that also has long gold summaries \cite{kryscinski-etal-2022-booksum}. 

We also show some generation examples in \Cref{tab:ex_qasper} and \ref{tab:ex_ssfd}.
We find that \ourschat{} can especially benefit from the additional contexts in the encoder in the QA datasets, where the answer may be localized to just one small part of the entire input.
On the other hand, summarization tasks do not catastrophically fail when the model only has access to only part of the input, as the model can still generate a coherent summary for the provided context, achieving reasonable ROUGE-L scores.

\begin{table*}[ht]
    \centering
    \small
    \begin{tabular}{>{\raggedright\arraybackslash\tt}p{0.95\textwidth}<{}}
        \toprule
            \vspace{-1em}
            \noindent \textbf{Encoder Input $C_1$:} \\
            We propose a novel pre-training method called BRLM, which can effectively alleviates the distance between different source language spaces.
            Our proposed approach significantly improves zero-shot translation performance, consistently surpassing pivoting and multilingual approaches. Meanwhile, the performance on supervised translation direction remains the same level or even better when using our method.
            Related Work
            In recent years, zero-shot translation in NMT has attracted widespread attention in academic research. Existing methods are mainly divided into four categories: pivot-based method, transfer learning, multilingual NMT, and unsupervised NMT.
            Pivot-based Method is a common strategy to obtain a source$\rightarrow $target model by introducing a pivot language. This approach is further divided into {\color{orange}pivoting and pivot-synthetic}. While the former firstly translates a source language into the pivot language which is later translated to the target language BIBREF4, BIBREF5, BIBREF12, the latter trains a source$\rightarrow $target model with pseudo \\\\
            
            \noindent \textbf{Encoder Input $C_2$:} \\
            , NMT heavily relies on large-scale parallel data, resulting in poor performance on low-resource or zero-resource language pairs BIBREF3. Translation between these low-resource languages (e.g., Arabic$\rightarrow $Spanish) is usually accomplished with pivoting through a rich-resource language (such as English), i.e., Arabic (source) sentence is translated to English (pivot) first which is later translated to Spanish (target) BIBREF4, BIBREF5. However, the pivot-based method requires doubled decoding time and suffers from the propagation of translation errors.
            One common alternative to avoid pivoting in NMT is transfer learning BIBREF6, BIBREF7, BIBREF8, BIBREF9 which leverages a high-resource pivot$\rightarrow $target model (parent) to initialize a low-resource source$\rightarrow $target model (child) that is further optimized with a small amount of available parallel data. Although this approach has achieved success in some low-resource language pairs, it still performs very poorly in extremely low-resource or zero-resource translation scenario. Specifically, BIBREF8 reports that without any child model training data, \\\\

            \noindent \textbf{Encoder Inputs $[C_3, \ldots, C_{17}]$ Omitted...} \\\\

            \noindent \textbf{Decoder Input $X$:} \\
            tokens are selected to be masked. Among the selected tokens, $80\%$ of them are replaced with [MASK] token, $10\%$ are replaced with a random BPE token, and $10\%$ unchanged. The prediction accuracy of masked words is used as a stopping criterion in the pre-training stage. Besides, we use fastalign tool BIBREF34 to extract word alignments for BRLM-HA.
            Experiments ::: Main Results
            Table TABREF19 and TABREF26 report zero-shot results on Europarl and Multi-UN evaluation sets, respectively. We compare our approaches with related approaches of pivoting, multilingual NMT (MNMT) BIBREF19, and cross-lingual transfer without pretraining BIBREF16. The results show that our approaches consistently outperform other approaches across languages and datasets, especially surpass pivoting, which is a strong baseline in the zero-shot scenario that multilingual NMT systems often fail to beat BIBREF19, BIBREF20, BIBREF23. Pivoting translates source to pivot then to target in two steps, causing inefficient translation process. Our approaches use one encoder-decoder model to translate between any zero-shot directions, which is more efficient than pivoting. Regarding the comparison between transfer approaches, our cross-lingual pretraining based transfer outperforms transfer method that does not use pretraining by a large margin.
            Experiments ::: Main Results ::: Results on Europarl Dataset.
            Regarding comparison between the baselines in table TABREF19, we find that pivoting is the strongest baseline that has significant advantage over other two baselines. Cross-lingual transfer for languages without shared vocabularies BIBREF16 manifests the worst performance because of not using source$\leftrightarrow $pivot parallel data, which is utilized as beneficial supervised signal for the other two baselines.
            \noindent \textbf{Additional Decoder Input Omitted...} \\

            You are given a scientific article and a question. Answer the question as concisely as you can, using a single phrase or sentence if possible. If the question cannot be answered based on the information in the article, write "unanswerable". If the question is a yes/no question, answer "yes", "no", or "unanswerable". \\
            
            Question: \\
            what are the pivot-based baselines? \\

            Answer:\\\\

            \noindent \textbf{Model Outputs:} \\
            \noindent {\color{blue} \llamachat{} output: unanswerable.}\\
            \noindent {\color{violet} \ours{} output with encoder contexts: pivot-based baselines include pivoting and pivot-synthetic.}\\

            \noindent {\color{orange} Gold answers: pivoting, pivoting$_{\rm m}$}\\
        \bottomrule
    \end{tabular}
    \caption{
        \zs{} generation example on the Qasper dataset.
        \ours{} sees the entire article through the decoder and the encoder, whereas \llamachat{} only sees a 2K token window.
        For brevity, we omit part of the decoder input and only show $2$ out of $k=17$ encoder inputs for \ours{}.
    }
    \label{tab:ex_qasper}
\end{table*}

\begin{table*}[ht]
    \centering
    \small
    \begin{tabular}{>{\raggedright\arraybackslash\tt}p{0.95\textwidth}<{}}
        \toprule
            \vspace{-1em}
            \noindent \textbf{Encoder Input $C_1$:} \\
            Phoebe: Almost sunrise. Do you think you're ready to try the window again?
            Prue: Yeah, yeah, but Abraxas will be ready for us here. We have to take him by surprise, go where we're most powerful, where we're most connected.
            [Cut to the park. Prue, Piper and Phoebe have joined hands around a stone.]
            Prue, Piper and Phoebe: "Hear now the words of the witches, the secrets we hid in the night, the oldest of Gods are invoked here, the great work of magic is sought."
            [Cut to Abraxas undoing the spell that gave them their powers.]
            [Cut back to the girls.]
            Prue, Piper, Phoebe: "In this night, and in this hour we call upon the ancient power."
            [Cut back to Abraxas. The pages of the Book of Shadows turn over. He turns back to the spell and continues to read it backwards.]
            [Cut back to the girls.]
            Prue, Piper, Phoebe: "Hear \\\\

            \noindent \textbf{Encoder Input $C_2$:} \\
            hoebe: Honey, we've done a lot of good as witches too. You know that. But that doesn't mean that bad things still aren't gonna happen. But just because we can't help that, doesn't mean that is our fault.
            Piper: Or yours.
            Phoebe: Come here.
            (They all hug.)
            Grams' Voice: The Power of Three.
            (They look around.)
            Phoebe: Okay, did you guys hear that?
            Piper: Mm-hmm.
            Prue: Grams?
            Grams' Voice: The Power of Three.
            Prue: How is that possible?
            Phoebe: I don't know but it's gotta to be a good thing if we're all hearing it.
            Piper: Maybe it means we're connected.
            Phoebe: To be as one. Maybe that's what the Book meant by the Power of One.
            Piper: The three of us working as one.
            Prue: Which is what the triquetra represents.
            (Phoebe looks at her watch) \\\\
            \noindent \textbf{Encoder Inputs $[C_3, \ldots, C_{44}]$ Omitted...} \\\\

            \noindent \textbf{Decoder Input $X$:} \\
            the secrets we hid in the night, the oldest of Gods are invoked here, the great work of magic is sought."
            [Cut back to Abraxas. The Book disappears.]
            [Cut back to the girls. The Book drops out from the sky and lands on the rock.]
            Phoebe: Prue, Piper...
            Prue: Keep chanting.
            Prue, Piper, Phoebe: Hear now the words of the witches, the secrets we hid in the night..." (Abraxas appears above them)
            Abraxas: What are you doing? How'd you bring me here?
            Prue, Piper, Phoebe: "The oldest of Gods are invoked here, the great work of magic is sought. In this night, and in this hour we call upon the ancient power. In this night, and in this hour we call upon the ancient power. In this night, and in this hour we call upon the ancient power."
            (Abraxas is vanquished. Phoebe grabs the Book.) 
            \noindent \textbf{Additional Decoder Input Omitted...} \\\\

            You are given a script of a TV episode. Summarize the episode in a paragraph. \\\\

            Summary:\\\\

            \noindent \textbf{Model Outputs:} \\
            \noindent {\color{blue} \llamachat{} output (ROUGE-L $= 19.1$): Prue, Piper, and Phoebe find an old Book of Shadows in the attic of their home, which they believe to be the original Book of Shadows. They start to read it, but Abraxas, the demon who wrote the book, appears to them, warning them to stop reading the book because he has returned to Earth and is seeking the power in the book to take over. The three sisters are able to defeat Abraxas and close the book, but the experience causes them to reflect on their lives as witches and their bond as sisters..}\\
            \noindent {\color{violet} \ours{} output with encoder contexts (ROUGE-L $= 19.3$): In this episode of Charmed, Prue is worried about the future because Andy is not coming back and they don't know what their new life will be like. Prue and Piper have a hard time coming to terms with the fact that their powers are no longer needed and they are losing their sense of purpose and identity. At the same time, they are struggling to cope with their feelings for each other and their relationship. Phoebe, meanwhile, has found a new source of power and is learning to control her abilities in a new way. As the three sisters work together to defeat an evil spirit that is threatening their home, they are able to come to terms with their powers and their relationship and find a new sense of purpose and identity.}\\

            \noindent {\color{orange} Gold answers: The sisters face a demon named Abraxas , who steals the Book of Shadows and undoes some of the spells they've cast (in reverse order), releasing old vanquished enemies of the Charmed Ones and forcing them to remember the vanquishing spells from memory alone. The sisters have one chance to recapture the Book of Shadows or they will lose their powers forever. The sisters also meet their new neighbors, Jenny and her uncle Dan Gordon . Phoebe and Piper learn through the Wiccan community that because their anniversary of activating their inherent powers falls on an equinox , a wiccan holy day , each of their powers will be more developed and greatly magnified but only temporarily.}\\
        \bottomrule
    \end{tabular}
    \caption{
        \zs{} generation example on the SummScreenFD dataset.
        \ours{} sees the entire TV script through the decoder and the encoder, whereas \llamachat{} only sees a 2K token window.
        For brevity, we omit part of the decoder input and only show $2$ out of $k=44$ encoder inputs for \ours{}.
    }
    \label{tab:ex_ssfd}
\end{table*}

\begin{table*}[ht]
    \centering
    \small
        \begin{tabular}{lrrrrrr}
            \toprule
            Dataset & Mean    & Median  & Max.    & Min.   & $l \in [4K, 32K]$ (\%) & $l > 32K$ (\%) \\
            \midrule
            NarrativeQA & 75998.7 & 69163   & 264143 & 18260 & 5.8                     & 94.2          \\
            QASPER & 5278.5  & 4896.0  & 17626  & 2003  & 61.6                    & 0.0           \\
            QuALITY   & 8035.8  & 8285.5  & 10779  & 3443  & 96.2                    & 0.0           \\
            GovReport & 11345.4 & 10128.5 & 48074  & 2441  & 91.2                    & 0.6           \\
            SummScreenFD & 9924.2  & 9364.0  & 27565  & 3078  & 99.1                    & 0.0           \\
            QMSum     & 16027.7 & 14811.5 & 34543  & 3120  & 94.1                    & 4.8           \\
            \bottomrule
        \end{tabular}
    \caption{
        \zs{} length statistics. $l$ is the number of input tokens in each example.
        We report the mean, median, maximum, and minimum number of input tokens in each dataset.
        We also report the percentage of examples that have between $4,096$ and $32,768$ tokens ($l \in [4K, 32K]$) and the percentage of examples that have over $32,768$ tokens ($l > 32K$).
        We observe that only NarrativeQA has a substantial number of test examples with more than 32K tokens, making it the most useful for evaluating long-context language models.
    }
    \label{tab:stats_zs}
\end{table*}

\section{Needle in the Haystack}
\label{app:needle}

\begin{figure}[t!]
    \centering
    \includegraphics[width=0.98\linewidth]{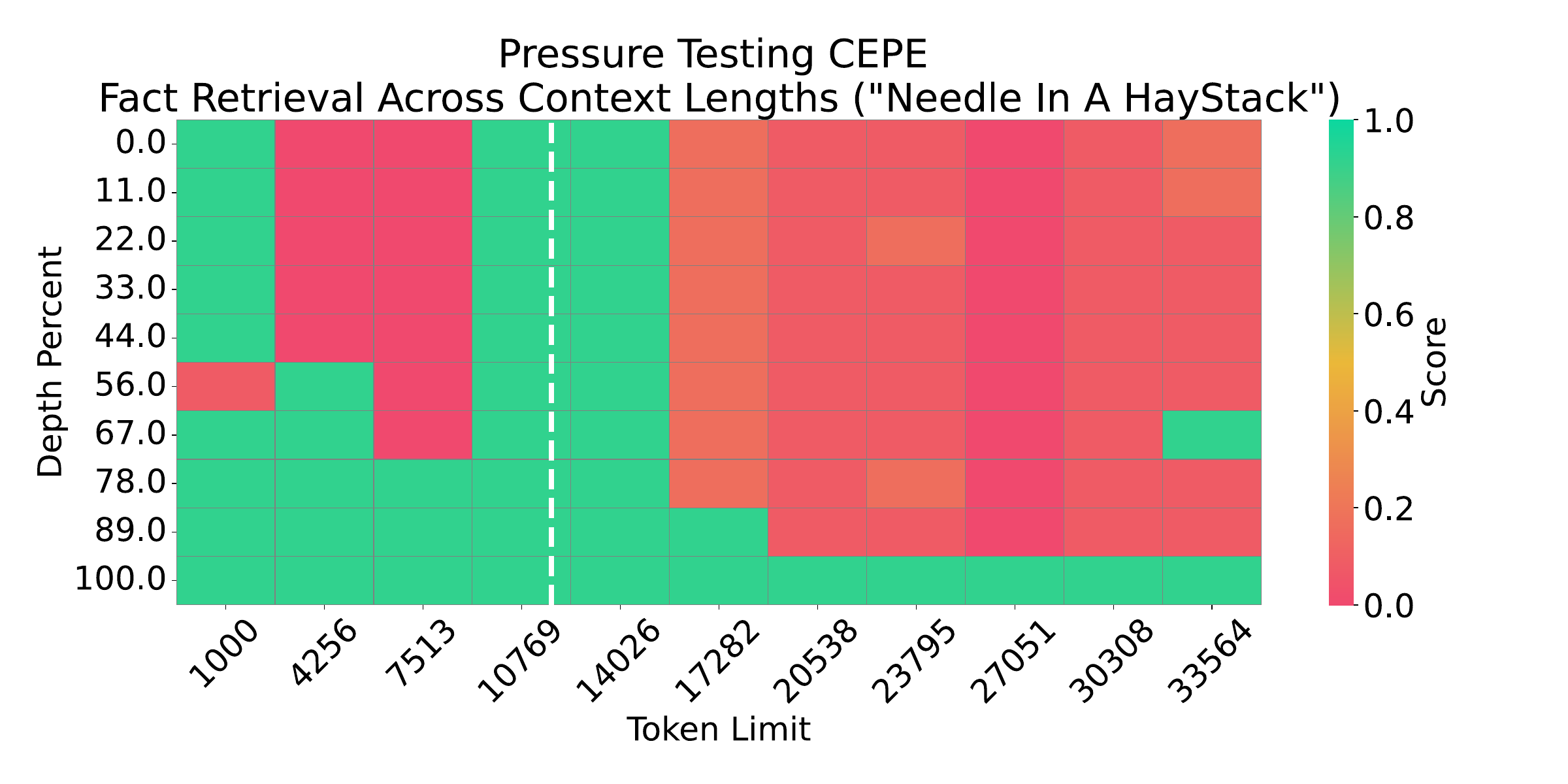}
    \caption{
        Needle in the Haystack evaluation on \oursllama.
        The white dotted line denotes the training length.
   }
    \vspace{-3pt}
    \label{fig:needle}
\end{figure}

Needle in the Haystack \cite{gkamradt_llmtest_needleinahaystack_2024} is a synthetic task designed to test \revision{in-context} retrieval abilities of long-context language models.
In this task, a needle (e.g., ``The best thing to do in San Francisco is to eat a sandwich and sit in Dolores Park on a sunny day'')
is placed in a long text of essays, and the model is tasked with retrieving the needle from the context---that is, the model is expected to answer ``What is the best thing to do in San Francisco?'' with the needle.

We evaluate \oursllama{}, where $2K$ tokens are used in the decoder, and the rest of the context is used in the encoder.
For this experiment, we also scale the cross-attention scores with the ratio between the number of tokens during inference and the number of tokens during training, following \citet{peng2024yarn}.
We follow the evaluation setting of \citet{fu2024data}, but split the context by sentence boundaries when inputting additional tokens to the encoder.
We find that while \oursllama{} is able to perfectly retrieve the needle in context at the training sequence length (\textasciitilde 10K tokens) and at some longer sequences (\textasciitilde 14K tokens), it struggles at other lengths.
\revision{This is likely due to the training/inference discrepancies on the input length and we will explore training objectives that can mitigate such discrepancies in the future.}
Other efficient methods, such as StreamingLLM~\cite{xiao2024efficient}, also struggle with copying tokens that are far from the current context.

\section{Ablations}
\label{app:ablation}

We show the full results for the ablation studies in \Cref{tab:ab_prevdoc_full} and \ref{tab:ab_retdoc_full}, and \ref{tab:ab_zs}.
The subsections below describe the ablation settings in more detail.

\subsection{Training Settings}
\label{app:ablations:training}

\paragraph{Training with retrieved documents.}
In this subsection, we study 
training \ours{} with retrieved documents.
Specifically, we pair the training sequences described in \S\ref{sec:methods:data} with retrieved passages from the retrieval split of RP using Contriever.
We use the first 256 tokens of the decoder input $X$ as the query and retrieve $k=16$ passages to form the additional contexts $\mathcal{C}$. 
Then, we train \ours{} with retrieved passages (\retdoc) using the same training settings as in \S\ref{sec:methods:training}.

\paragraph{Encoder training.}
We also investigate how to best train the encoder.
To this end, we train \ours{} with (1) freezing the encoder after pre-training and the warmup stage, (2) training with a randomly initialized encoder, and (3) using the pre-trained model without the warmup stage.
As shown in Table \ref{tab:ab_retdoc}, we find that 
the copying warmup stage 
and fine-tuning the encoder during training are both crucial for strong performance.

\subsection{KL Divergence}
\label{app:ablations:kl}
The key component of \ourschat{} is the KL Divergence loss.
To understand the importance of this auxiliary loss, we explore the performance of \ourschat{} with changes to the loss function in this subsection.
Let $\mathcal{L}_{CE}, \mathcal{L}_{KL}$ be the cross entropy loss and the KL Divergence loss, respectively.
Then, the total loss is $\mathcal{L} = c_{CE}\mathcal{L}_{CE} + c_{KL} \mathcal{L}_{KL}$, where $c_{CE}$ and $c_{KL}$ are coefficients for the cross entropy loss and the KL Divergence loss, respectively.
We vary the coefficients $c_{CE}$ and $c_{KL}$ to study the importance of the KL Divergence loss.
The results are presented in Table \ref{tab:ab_zs}.
Without the KL Divergence loss, \ourschat{} may still perform well on NarrativeQA and Qasper, but the performance on the summarization tasks and QuALITY may decrease as a result.

\begin{table*}[t]
    \centering
    \small
    \begin{tabular}{lcccccc}
        \toprule
        & \textbf{ArXiv}           & \textbf{Book}          & \textbf{PG19}     & \textbf{ProofPile} & \textbf{CodeParrot}\\
        \midrule
        \multicolumn{3}{l}{\textbf{Total Tokens $= 4096$}} &       &       &       \\
        \midrule
        \ours                   & 2.579       & 6.292      & 7.536 & 2.396 & 1.763 \\
        w/ \retdoc              & 2.649       & 6.340      & 7.586 & 2.465 & 1.775 \\
        w/ RP Only              & 2.633       & 6.335      & 7.604 & 2.446 & 1.766 \\
        w/ AB Only              & 2.569       & 6.287      & 7.525 & 2.386 & 1.772 \\
        w/ Frozen Encoder       & 2.631       & 6.353      & 7.603 & 2.446 & 1.785 \\
        w/ Random Encoder       & 2.680       & 6.374      & 7.617 & 2.488 & 1.797 \\
        w/ No Warmup         & 2.678       & 6.372      & 7.613 & 2.487 & 1.796 \\
        \midrule
        \multicolumn{5}{l}{\textbf{Total Tokens $=  8192$}}                &       \\
        \midrule
        \ours                   & 2.496       & 6.049      & 7.372 & 2.219 & 1.715 \\
        w/ \retdoc              & 2.553       & 6.089      & 7.417 & 2.278 & 1.724 \\
        w/ RP Only              & 2.543       & 6.085      & 7.434 & 2.262 & 1.718 \\
        w/ AB Only              & 2.485       & 6.040      & 7.357 & 2.208 & 1.720 \\
        w/ Frozen Encoder       & 2.541       & 6.099      & 7.430 & 2.261 & 1.734 \\
        w/ Random Encoder       & 2.571       & 6.108      & 7.439 & 2.291 & 1.739 \\
        w/ No Warmup         & 2.572       & 6.113      & 7.439 & 2.292 & 1.739 \\
        \midrule
        \multicolumn{5}{l}{\textbf{Total Tokens $= 32768$}}                &       \\
        \midrule
        \ours                   & 2.421       & 6.015      & 7.204 & 2.218 & 1.702 \\
        w/ \retdoc              & 2.546       & 6.088      & 7.280 & 2.332 & 1.726 \\
        w/ RP Only              & 2.497       & 6.059      & 7.271 & 2.288 & 1.709 \\
        w/ AB Only              & 2.396       & 5.995      & 7.178 & 2.195 & 1.702 \\
        w/ Frozen Encoder       & 2.520       & 6.091      & 7.282 & 2.297 & 1.739 \\
        w/ Random Encoder       & 2.571       & 6.108      & 7.303 & 2.346 & 1.752 \\
        w/ No Warmup         & 2.571       & 6.110      & 7.301 & 2.346 & 1.752 \\
        \bottomrule
    \end{tabular}
    \caption{
        Test perplexity for all ablation settings in the long-context language modeling evaluation setting.
    }
    \vspace{-2pt}
    \label{tab:ab_prevdoc_full}
\end{table*}

\begin{table*}[t]
    \centering
    \small
    \begin{tabular}{lcccccccc}
        \toprule
        & \textbf{ArXiv} & \textbf{Book} & \textbf{C4-RP} & \textbf{CC} & \textbf{Github} & \textbf{StackEx} & \textbf{Wiki} & \textbf{Avg.}\\
        \midrule
        \multicolumn{7}{l}{\textbf{Total Tokens $= 2048$} ($k=0$)} \\
        \midrule
        \ours             & 3.486 & 6.481 & 6.884 & 5.319 & 1.793 & 3.709 & 4.302 & 4.568 \\
        w/ \retdoc        & 3.413 & 6.399 & 6.854 & 5.263 & 1.788 & 3.694 & 4.287 & 4.528 \\
        w/ RP Only        & 3.485 & 6.479 & 6.901 & 5.313 & 1.777 & 3.700 & 4.281 & 4.562 \\
        w/ AB Only        & 3.505 & 6.504 & 7.185 & 5.444 & 1.859 & 4.018 & 4.763 & 4.754 \\
        w/ Frozen Encoder & 3.501 & 6.495 & 6.933 & 5.505 & 1.821 & 3.734 & 4.323 & 4.616 \\
        w/ Random Encoder & 3.426 & 6.442 & 6.904 & 5.541 & 1.838 & 3.728 & 4.338 & 4.602 \\
        w/ No Copy Init   & 3.452 & 6.459 & 6.914 & 5.546 & 1.842 & 3.732 & 4.344 & 4.613 \\
        \midrule
        \multicolumn{8}{l}{\textbf{Total Tokens $= 7168$} ($k=20$)}               &       \\
        \midrule
        \ours             & 3.475 & 6.463 & 6.875 & 5.266 & 1.782 & 3.703 & 4.296 & 4.551 \\
        w/ \retdoc        & 3.413 & 6.393 & 6.839 & 5.169 & 1.779 & 3.693 & 4.286 & 4.510 \\
        w/ RP Only        & 3.479 & 6.467 & 6.894 & 5.249 & 1.767 & 3.696 & 4.276 & 4.547 \\
        w/ AB Only        & 3.491 & 6.481 & 7.140 & 5.401 & 1.846 & 4.004 & 4.738 & 4.729 \\
        w/ Frozen Encoder & 3.485 & 6.482 & 6.930 & 5.500 & 1.815 & 3.727 & 4.318 & 4.608 \\
        w/ Random Encoder & 3.426 & 6.442 & 6.904 & 5.540 & 1.837 & 3.727 & 4.337 & 4.602 \\
        w/ No Copy Init   & 3.447 & 6.457 & 6.913 & 5.545 & 1.841 & 3.721 & 4.332 & 4.608 \\
        \midrule
        \multicolumn{8}{l}{\textbf{Total Tokens $= 14848$} ($k=50$)}              &       \\
        \midrule
        \ours             & 3.467 & 6.457 & 6.881 & 5.273 & 1.777 & 3.701 & 4.292 & 4.550 \\
        w/ \retdoc        & 3.413 & 6.392 & 6.835 & 5.098 & 1.776 & 3.692 & 4.287 & 4.499 \\
        w/ RP Only        & 3.472 & 6.465 & 6.900 & 5.243 & 1.762 & 3.693 & 4.274 & 4.544 \\
        w/ AB Only        & 3.480 & 6.471 & 7.114 & 5.412 & 1.838 & 3.994 & 4.719 & 4.718 \\
        w/ Frozen Encoder & 3.474 & 6.474 & 6.930 & 5.509 & 1.814 & 3.723 & 4.316 & 4.606 \\
        w/ Random Encoder & 3.426 & 6.442 & 6.904 & 5.540 & 1.837 & 3.726 & 4.337 & 4.602 \\
        w/ No Copy Init   & 3.445 & 6.456 & 6.913 & 5.545 & 1.841 & 3.716 & 4.329 & 4.606 \\
        \bottomrule
    \end{tabular}
    \caption{
        Test perplexity on RedPajama across all domains in the retrieval-augmented setting for all ablation experiments.
        $k$ is the number of additional contexts used.
        Avg. is the macro average across all domains. 
    }
    \vspace{-2pt}
    \label{tab:ab_retdoc_full}
\end{table*}

\begin{table}[th]
    \centering
    \resizebox{0.98\linewidth}{!}{
        \begin{tabular}{llcccccc}
            \toprule
             & \multicolumn{3}{c}{\textbf{Question Answering}} & \multicolumn{3}{c}{\textbf{Summarization}} \\ 
            \cmidrule(lr){2-4} \cmidrule(lr){5-7}
            $c_{KL}$ & NQA & Qspr & QALT & GvRp & SSFD  & QMSum \\
            \midrule
            \multicolumn{7}{l}{\textbf{Total Tokens} $= 4$K}   \\
            \midrule
            2        & 19.5 & 20.5 & \textbf{30.2} & \textbf{16.5} & \textbf{16.4} & \textbf{19.6} \\
            1        & \textbf{21.6} & 20.7 & 27.2 & 16.3 & 5.3  & 4.7  \\
            0        & 21.3 & 21.0 & \textbf{27.4} & 14.6 & 14.9 & 15.6 \\
            \midrule
            \multicolumn{7}{l}{\textbf{Total Tokens} $= 16$K}  \\
            \midrule
            2        & 20.6 & 19.9 & \textbf{29.6} & 15.9 & \textbf{16.8} & \textbf{19.4} \\
            1        & 21.8 & \textbf{20.6} & 26.8 & \textbf{16.0} & 15.2 & 16.1 \\
            0        & \textbf{22.9} & \textbf{20.6} & 26.4 & 14.8 & 5.2  & 4.7  \\
            \midrule
            \multicolumn{7}{l}{\textbf{Total Tokens} $= 32$K}  \\
            \midrule
            2        & 21.6 & 19.9 & \textbf{29.6} & 15.8 & \textbf{16.7} & \textbf{19.5} \\
            1        & \textbf{22.7} & \textbf{20.6} & 26.8 & \textbf{16.0} & 15.2 & 15.8 \\
            0        & 22.3 & \textbf{20.6} & 26.4 & 14.6 & 5.2  & 4.7  \\
            \midrule
            \multicolumn{7}{l}{\textbf{All Tokens}}            \\
            \midrule
            2        & 21.9 & 19.9 & \textbf{29.6} & \textbf{15.9} & \textbf{16.7} & \textbf{19.5} \\
            1        & 22.6 & \textbf{20.6} & 26.8 & \textbf{15.9} & 15.2 & 15.8 \\
            0        & \textbf{23.0} & \textbf{20.6} & 26.4 & 14.6 & 5.2  & 4.7  \\
            \bottomrule
       \end{tabular}
    }
    \caption{
        \zs{} results using different losses during training, 
        where 
        $c_{KL}$ is the coefficient of the KL Divergence loss.
        $c_{CE} = 1$ is the coefficient of the Cross-Entropy loss for all experiments.
    }
    \label{tab:ab_zs}
\end{table}

\end{document}